\documentclass{article}

\usepackage{arxiv}

\usepackage[utf8]{inputenc} 
\usepackage[T1]{fontenc}    
\usepackage{hyperref}       
\usepackage{url}            
\usepackage{booktabs}       
\usepackage{amsfonts,amsmath,amsthm,amssymb} 
\usepackage{nicefrac}       
\usepackage{microtype}      
\usepackage{graphicx,subfig,color,xcolor} 
\usepackage{float}          
\usepackage{multirow}       
\usepackage{tabularx}
\usepackage{ulem}
\usepackage{algorithm}      
\usepackage{algpseudocode}  
\usepackage{mathrsfs}

\usepackage{graphicx}
\usepackage{amssymb}
\usepackage{amsmath}
\usepackage{amsthm}
\usepackage{amsmath}
\usepackage{xcolor}
\usepackage{caption}
\usepackage{subcaption}
\usepackage{mathrsfs}

\theoremstyle{remark}
\newtheorem{remark}{Remark}

\title{Morephy-Net: An Evolutionary Multi-objective Optimization for Replica-Exchange-based Physics-informed Neural Operator Learning Networks}

\author{
  Binghang Lu \\
  School of Electrical and Computer Engineering\\
  Purdue University\\
  West Lafayette, IN 47907, USA \\
  \And
  Changhong Mou \\
  Department of Mathematics and Statistics\\
  Utah State University\\
  Logan, UT 84322, USA \\
  \And
  Guang Lin\thanks{Corresponding author} \\
  Department of Mathematics \& School of Mechanical Engineering\\
  Purdue University\\
  West Lafayette, IN 47907, USA \\
  \texttt{guanglin@purdue.edu} \\
}

\begin{document}
\maketitle

\begin{abstract}
  We propose an evolutionary \textbf{M}ulti-objective \textbf{O}ptimization for \textbf{R}eplica-\textbf{E}xchange-based \textbf{Phy}sics-informed operator-learning \textbf{Net}works (Morephy-Net) to solve parametric partial differential equations (PDEs) in noisy data regimes, for both forward prediction and inverse identification. Existing physics-informed neural networks and operator-learning models (e.g., DeepONets and Fourier neural operators) often face three coupled challenges: (i) balancing data/operator and physics residual losses, (ii) maintaining robustness under noisy or sparse observations, and (iii) providing reliable uncertainty quantification. Morephy-Net addresses these issues by integrating: (i) evolutionary multi-objective optimization that treats data/operator and physics residual terms as separate objectives and searches the Pareto front, thereby avoiding ad hoc loss weighting; (ii) replica-exchange stochastic gradient Langevin dynamics to enhance global exploration and stabilize training in non-convex landscapes; and (iii) Bayesian uncertainty quantification obtained from stochastic sampling. We validate Morephy-Net on representative forward and inverse problems, including the one-dimensional Burgers equation and the time-fractional mixed diffusion--wave equation. The results demonstrate consistent improvements in accuracy, noise robustness, and calibrated uncertainty estimates over standard operator-learning baselines.
\end{abstract}


\section{Introduction}
The accurate and efficient solution of partial differential equations (PDEs) is indispensable in modeling and understanding a wide range of physical, biological, and engineering systems \cite{evans2022partial,strauss2007partial,farlow1993partial}. Solving parametric PDEs with varying geometries, parameters, and initial/boundary conditions is equivalent to learning the solution operator that maps inputs to their corresponding solutions \cite{hughes2003finite,leveque2007finite, hamming2012numerical}.
For complex systems, traditional methods such as the finite difference method (FDM) and finite element method (FEM) often require prohibitive computational resources as they need fine spatial–temporal discretization and repeated simulations for different parameters and initial/boundary conditions \cite{szabo2021finite,layton2008introduction,berselli2006mathematics,strikwerda2004finite,leveque2007finite}. 
In recent years, machine learning (ML) has emerged as a powerful and increasingly popular alternative to traditional numerical solvers for parametric PDEs \cite{brunton2024promising,raissi2018hidden,lu2021learning,mou2023combining,kharazmi2021identifiability,gu2023stationary,teng2023level,manti2025symbolic}.
There are several different ML approaches for solving PDEs. Among them, one prominent category is \textit{physics-informed neural networks (PINNs)} \cite{raissi2019physics}, which incorporate the underlying physical laws directly into the training process.
PINNs have shown success in approximating solutions to specific PDE instances and have been extended to learn solution operators that generalize across varying inputs (i.e., parametric PDEs). However, training PINNs for parametric PDEs remains computationally expensive, particularly in multi-query or real-time scenarios, and this setting often requires overly complex parameterizations \cite{wang2022respecting,krishnapriyan2021characterizing}.
Another alternative is \textit{operator learning}, such as the Deep Operator Network (DeepONet or DON) \cite{lu2021learning} and Fourier Neural Operator (FNO) \cite{li2021fourier}, which aim to directly learn the mapping from input functions to PDE solutions. However, their performance often deteriorates in challenging settings, including inverse problems, noisy or sparse observational data, and high-dimensional parameter spaces \cite{kovachki2024data}.

To address the limitations of traditional operator learning, Wang et al. \cite{wang2021learning} proposed a physics-informed DeepONet framework for parametric PDEs that combines the operator-learning capability of DeepONets with PDE residual constraints, as in PINNs. 
This framework reduces the dependence of general operator learning on large amounts of paired high-resolution training data and preserves consistency to the original PDEs with different discretizations. In particular, physics-informed DeepONets adopt a composite loss function consisting of: (1) an operator loss, and (2) a physics loss derived from the PDE residual.  \cite{wang2021learning} showed that physics-informed DeepONets are data-efficient and particularly effective in small-data regimes. Nevertheless, several challenges inherent to the physics-informed DeepONets framework persist. First, similar to PINNs, the training of physics-informed DeepONets admits a multi-objective optimization problem, where the operator loss and the physics loss from the PDE residual must be balanced appropriately \cite{lu2023nsga,lu2025mopinnenkf}. Second, in real applications, the available data are often contaminated with noise. Such noise can diminish the accuracy of the learned operator and even destabilize the training process, particularly in inverse problem settings where the physics loss alone may be insufficient to adequately regularize the solution \cite{kovachki2024data,ding2023learning}. Third, since operator learning optimization often relies on gradient-based methods such as the Adam optimizer, the current physics-informed DeepONet framework lacks the ability for uncertainty quantification \cite{winovich2025active,zou2024neuraluq}.

To address these challenges, we propose a new \textit{\textbf{M}ulti-\textbf{O}bjective \textbf{R}eplica-\textbf{E}xchange \textbf{Phy}sics-\textbf{I}nformed neural-operator \textbf{Net}work} (Morephy-Net). The novelty of Morephy-Net is to unify (a) principled multi-objective training, (b) replica-exchange posterior sampling, and (c) uncertainty quantification within a single physics-informed operator-learning framework. Concretely, Morephy-Net introduces:
(i) an evolutionary multi-objective optimizer that treats the operator/data loss and physics residual loss as separate objectives and searches for Pareto-optimal trade-offs, eliminating manual loss weighting;
(ii) replica-exchange stochastic gradient Langevin dynamics (reSGLD) as an efficient sampler to enhance global exploration and reduce the risk of converging to poor local minima; and
(iii) built-in uncertainty quantification (UQ) based on the resulting stochastic samples, enabling uncertainty-aware predictions for both forward and inverse problems.
Morephy-Net adopts a modified reference-point-based non-dominated sorting method (refined NSGA-III) \cite{deb2002fast} and further improves diversity along the Pareto front via the proposed refined Pareto sampling (RPS) strategy. By promoting diverse Pareto solutions during training, RPS mitigates Pareto-front clustering and yields more stable trade-offs between data fidelity and physical consistency.
The reSGLD component is designed to improve parameter-space exploration. Standard stochastic gradient Langevin dynamics (SGLD)~\cite{welling2011bayesian} augments stochastic gradient descent with Gaussian noise to approximate posterior sampling during optimization; however, in high-dimensional non-convex landscapes it can still become trapped in local minima. In such cases, reSGLD~\cite{deng2020non,deng2022interacting,ZhengD00L24,liang2025bayesian} runs multiple SGLD processes at different temperatures and periodically exchanges their states. High-temperature chains explore broader regions of the energy landscape and facilitate escape from local traps, while low-temperature chains focus on fine-tuning around promising optima.
This mechanism enables Morephy-Net to converge faster, deliver more reliable parameter estimates, and quantify uncertainty for both forward and inverse PDE problems. We summarize the main contributions of this work as follows:
\vspace*{-.1cm}
\begin{enumerate}
    \item Morephy-Net incorporates \textit{evolutionary multi-objective optimization} into physics-informed operator learning via NSGA-III, treating the data and operator loss and the physics residual loss as distinct objectives and thereby avoiding manual loss weighting.
    \item We propose a refined Pareto sampling (RPS) strategy within NSGA-III to increase sampling diversity by re-selecting distinctive solutions along the Pareto front and mitigating local clustering.
    \item We employ \textit{replica-exchange stochastic gradient Langevin dynamics (reSGLD)} to enhance global exploration of the parameter space: high-temperature chains traverse broader energy landscapes to escape local minima, while low-temperature chains refine solutions; together, this accelerates convergence and improves accuracy.
    \item The stochastic sampling in reSGLD naturally enables \textit{uncertainty quantification (UQ)} for physics-informed operator learning in both forward and inverse PDE settings.
\end{enumerate}
\vspace*{-.1cm}
The remainder of this paper is organized as follows: Section~\ref{sec:general} describes the proposed Morephy-Net framework in detail that includes its main components: physics-informed DeepONet (Section~\ref{subsec:pi-deeponet}), NSGA-III multi-objective optimization (Section~\ref{subsec:nsga3}), Replica Exchange Stochastic gradient Langevin dynamics (Section~\ref{subsec:sgld-re}), and its uncertainty quantification (Section~\ref{subsec:uq}). Section~\ref{sec:numeric} presents numerical experiments on different test problems. Finally, Section~\ref{sec:conclusion} concludes the paper and discusses potential future directions.
\raggedbottom

\section{General framework \label{sec:general}} 
\subsection{Overview}

\begin{figure}[H]
    \centering
    \includegraphics[width=1.1\textwidth]{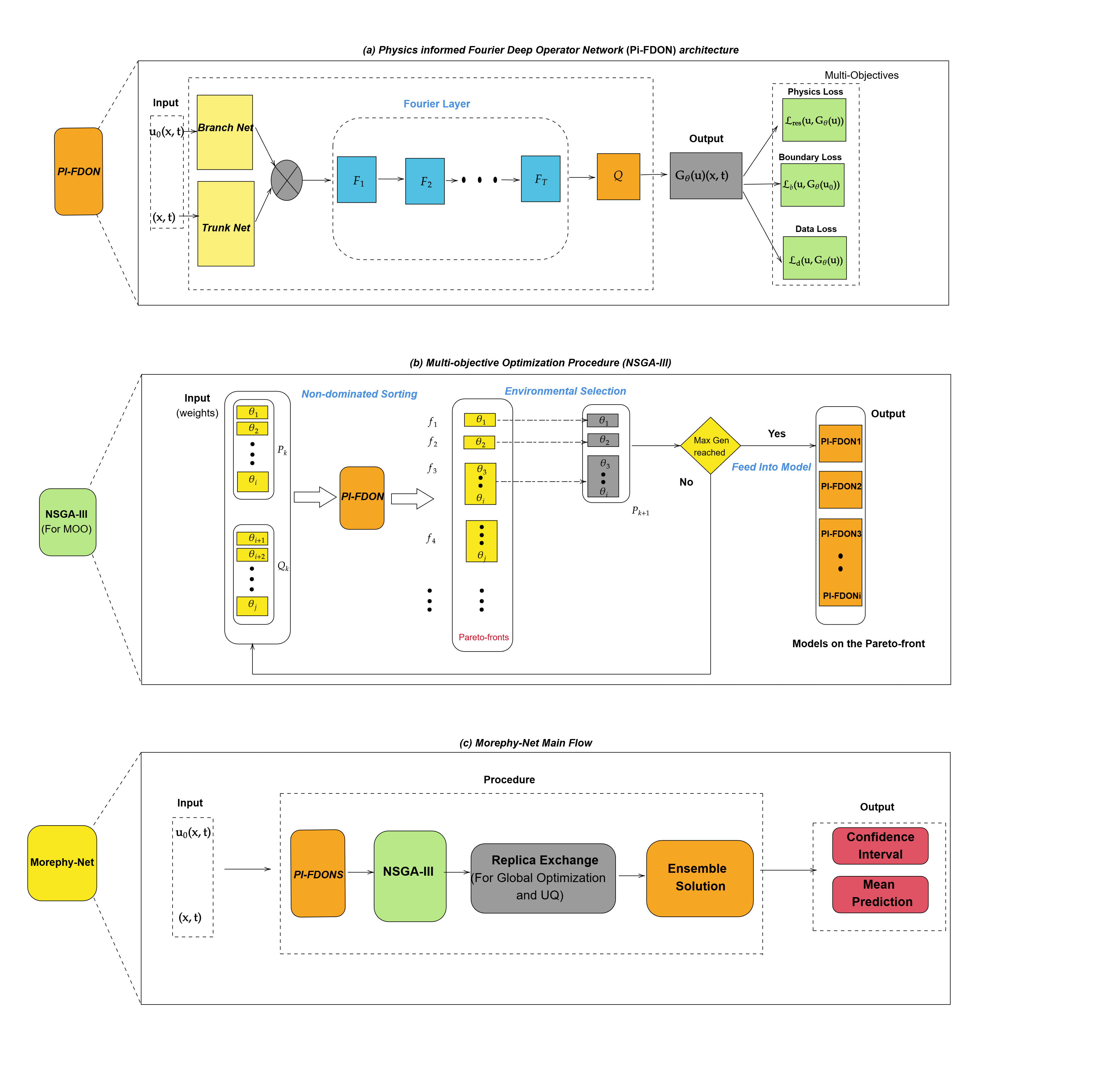}
    \caption{Schematic illustration of Morephy-Net framework.
   Top panel (a): The Physics-informed Fourier Deep Operator Network (PI-FDON) architecture consists of branch and trunk subnetworks that encode PDE parameters/initial conditions and spatio-temporal inputs, respectively. Training is guided by both data loss and physics-informed residual loss, with a Fourier layer applied after the dot product of the branch and trunk networks. Middle panel (b) Multi-objective optimization with the refined Non-dominated Sorting Genetic Algorithm III (refined NSGA-III). 
   Bottom panel (c). The main flow of the Morephy-Net. Replica exchange stochastic gradient Langevin dynamics (reSGLD) with variance reduction is used for uncertainty quantification. 
\label{fig:workflow}
     }
\end{figure}

The \textbf{M}ulti-objective \textbf{O}ptimization for \textbf{R}eplica-\textbf{E}xchange-based \textbf{Phy}sics-informed operator-learning \textbf{Net}works (Morephy-Net)framework consists of several key components, as illustrated schematically in Figure~\ref{fig:workflow}. In particular, Morephy-Net consists of the following:
(1) \textit{\textbf{Physics-informed Deep Operator Network (PI-DON) architecture.}}   The Deep Operator Network employs two subnetworks—the branch network that includes the initial conditions or PDE parameters, and the trunk network, which may contain the spatial temporal grid information. The training is guided not only by data loss but also by a physics-informed residual loss that enforces the governing PDE constraints (see Section \ref{subsec:pi-deeponet}).
(2) \textit{\textbf{Fourier convolution layer.}} The Fourier convolution layer maps the latent representation into a spectral space, where the underlying information can be more effectively represented by the neural network. This transformation also allows the model to capture global dependency and long-range interactions inherently in data.
(3) \textit{\textbf{Multi-objective optimization with replica exchange stochastic gradient Langevin dynamics.}} In Morephy-Net, the training process integrates the refined Non-dominated Sorting Genetic Algorithm III (refined NSGA-III) algorithm (see Section \ref{subsec:nsga3}), proposed in this paper, with replica exchange stochastic gradient Langevin dynamics (reSGLD) (see Section \ref{subsec:sgld-re}) to address the challenges of multi-objective optimization in physics-informed operator learning. The refined NSGA-III serves as the evolutionary mechanism for exploring the Pareto front that produces optimal candidate solutions. These candidates are then further optimized through reSGLD, which improves exploration of the parameter space by including stochastic gradient noise and enabling replica exchanges across parallel chains at different “temperatures.” This stochastic strategy not only helps the models escape local minima but also improves convergence in training.

\subsection{Physics Informed Deep Operator Network} \label{subsec:pi-deeponet}
The general deep operator network (DeepONet) proposed \cite{lu2021learning} with mathematical foundation of the universal operator approximation theorem \cite{chen1995universal} is intended to learn nonlinear operator mappings between infinite-dimensional function spaces.
For a general PDE problem, we introduce the solution operator  
\begin{align}
\mathcal{G}(\kappa) = u(\cdot;\kappa),    
\end{align}
where $\kappa$ represents the input, which can be the initial condition or parameters of the PDE. A DeepONet is a class of neural network models designed to approximate $\mathcal{G}$.
In particular, we denote by $G_\theta$ the DeepONet approximation of $\mathcal{G}$, with $\theta$ representing the trainable parameters of the network. A DeepONet consists of two subnetworks---the branch net and the trunk net---and the estimator evaluated at $\mathbf{x} \in M$ is an inner product of the branch and trunk outputs:
\begin{align}
G_\theta(\kappa(\Xi))(\mathbf{x}) = \sum_{k=1}^p b_k(\kappa(\xi_1), \kappa(\xi_2), \ldots, \kappa(\xi_m)) \, t_k(\mathbf{x}) + b_0,    
\end{align}
where $b_0 \in \mathbb{R}$ is a bias, $\{b_1, b_2, \ldots, b_p\}$ are the $p$ outputs of the branch net, and $\{t_1, t_2, \ldots, t_p\}$ are the $p$ outputs of the trunk net. 
More specifically, the branch network represents $\kappa$ in a discrete format: 
\begin{align}
\kappa(\Xi) = \{\kappa(\xi_1), \kappa(\xi_2), \ldots, \kappa(\xi_m)\}, 
\end{align}
where $\kappa(\Xi)$ is a vector that consists of evaluations of the input function at sensor locations $\Xi = \{\xi_1, \xi_2, \ldots, \xi_m\}$.
The trunk network  may take as input a spatial location $\mathbf{x} \in M$. 
Our goal is to predict the solution corresponding to a given $\kappa$, which is also evaluated at the same input location $\mathbf{x} \in M$. 
The representation of $\kappa$ through pointwise evaluations at arbitrary sensor locations provides additional flexibility, both during training and in prediction, particularly when $\kappa$ is available only through its values at these sensor points.

\begin{figure}[H]
\centering
\includegraphics[width=1.0\linewidth]{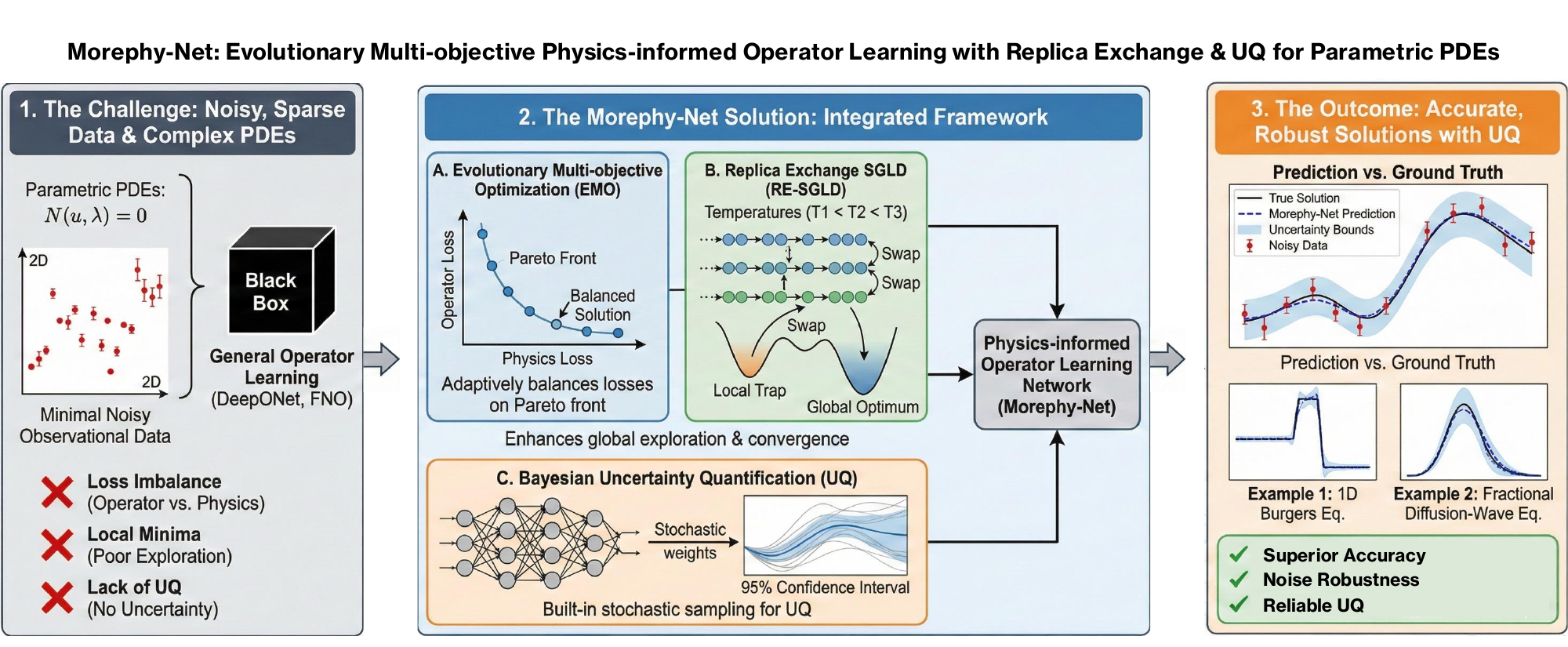}
\caption{Diagram of Morephy-Net. The framework integrates three key components: (A) evolutionary multi-objective optimization (EMO), (B) replica exchange stochastic gradient Langevin dynamics (RE-SGLD), and (C) Bayesian uncertainty quantification (UQ).}
\label{fig:morephy-net}
\end{figure}

Figure~\ref{fig:morephy-net} outlines the Morephy-Net approach for learning solution operators from noisy and sparse data. The framework addresses common limitations of conventional methods (e.g., loss imbalance and poor exploration) by integrating three components: (A) evolutionary multi-objective optimization (EMO) to adaptively balance data/operator and physics losses along the Pareto front; (B) replica exchange stochastic gradient Langevin dynamics (RE-SGLD) to escape local minima and improve global exploration; and (C) Bayesian uncertainty quantification (UQ) via stochastic sampling. As demonstrated on benchmark problems such as the 1D Burgers and fractional diffusion--wave equations, Morephy-Net yields a robust operator-learning model with improved accuracy, stronger noise resilience, and reliable uncertainty estimates.

Given the training data with labels in the two sets of inputs and outputs:
\begin{align}
&  \text{input:}  &\Big\{\kappa^{(k)}, \Xi, (\mathbf{x}_i^{(k)})_{i=1,\ldots,N}\Big\}_{k=1,\ldots,N_{\text{data}}}
    \\
 &   \text{output:}
&\{u(\mathbf{x}_i^{(k)}; \kappa^{(k)})\}_{i=1,\ldots,N,\; k=1,\ldots,N_{\text{data}}}.
\end{align}
we minimize a loss function that measures the discrepancy between the true solution operator $G(\kappa^{(k)})$ and its DeepONet approximation $G_\theta(\kappa^{(k)})$:
\begin{align}
\widehat{\mathcal{L}}_{\text{data}}(\theta) 
= \frac{1}{N_{\text{data}}N} \sum_{k=1}^{N_{\text{data}}} \sum_{i=1}^N 
\Big| G_\theta(\kappa^{(k)}(\Xi))(\mathbf{x}_i^{(k)}) 
     - G(\kappa^{(k)})(\mathbf{x}_i^{(k)}) \Big|^2,
\end{align}
where $X^{(k)} = \{\mathbf{x}_1^{(k)}, \ldots, \mathbf{x}_N^{(k)}\}$ denotes the set of $N$ evaluation points in the domain of $G(\kappa^{(k)})$.  
In practice, the analytical solution of the PDE is usually unavailable. Instead, we rely on numerical simulations from high-fidelity solvers, denoted by
\begin{align}
\big\{\hat{u}(\mathbf{x}_i^{(k)}; \kappa^{(k)}) \big\}_{i=1,\ldots,N,\; k=1,\ldots,N_{\text{data}}}.    
\end{align}
Accordingly, in practice we minimize the loss function
\begin{align}
{\mathcal{L}}_{\text{data}}(\theta) 
= \frac{1}{N_{\text{data}}N} \sum_{k=1}^{N_{\text{data}}} \sum_{i=1}^N 
\Big| G_\theta(\kappa^{(k)}(\Xi))(\mathbf{x}_i^{(k)}) 
     - \hat{u}(\mathbf{x}_i^{(k)}; \kappa^{(k)}) \Big|^2.
\end{align}

The Physics-Informed Deep Operator Network (PI-DeepONet or PI-DON; referred to as PI-DON in this paper) \cite{wang2021learning} integrates the idea of physics-informed neural networks (PINN) \cite{raissi2019physics} with the DeepONet framework. 
The PI-DON introduces the additional PDE residual loss function and boundary condition loss function.
Consequently, the loss function for PI-DON is defined as
\begin{align}
\mathcal{L}(\theta) 
= w_{\text{data}} \mathcal{L}_{\text{data}}(\theta) 
+ w_{\text{PDE}} \mathcal{L}_{\text{PDE}}(\theta) 
+ w_{\text{bc}} \mathcal{L}_{\text{bc}}(\theta),
\label{loss-pi-don}
\end{align}
where $\mathcal{L}_{\text{PDE}}(\theta)$ denotes the PDE residual loss, $\mathcal{L}_{\text{bc}}(\theta)$ the boundary condition loss, and $w_{\text{data}}, w_{\text{PDE}},$ and $w_{\text{bc}}$ are the corresponding weights for each term.

\subsection{Multi-objective Optimization} \label{subsec:nsga3}
To obtain an optimal PI-DON, i.e., to minimize the composite loss in \eqref{loss-pi-don}, we must solve a multi-objective optimization problem.
Multi-objective optimization problems frequently arise in scientific and engineering applications \cite{tian2021evolutionary,hong2021evolutionary,li2019quality,rockendorf2022design,park2024techno,pan2022evolutionary,wei2024bayesian,singh2025multi}, many of which involve high-dimensional search spaces.
In this paper, we propose a refined version of the nondominated sorting genetic algorithm (refined NSGA-III) that builds upon the classical NSGA-III framework originally introduced by \cite{deb2002fast}. The refined NSGA-III retains the population-based evolutionary structure designed for multi-objective optimization while incorporating additional mechanisms to improve sampling diversity in the operator learning settings.

\subsubsection{Nondominated Sorting Genetic Algorithm (NSGA-III)}
To illustrate the NSGA-III algorithm, we begin with the following general setting. Given $m\in \mathbf{N}$, an \emph{$m$-objective function} is defined as $f(x)=(f_1(x),\ldots, f_m(x))$, where $x\in\Omega$ and $f_i\colon \Omega\rightarrow \mathbf{R}$ for a given search space~$\Omega$.
Unlike single-objective optimization, there is usually no single solution that minimizes all $m$ objectives simultaneously. For comparison, we say that a solution $x$ \emph{dominates} a solution $y$, denoted $x\preceq y$, if and only if $f_j(x)\le f_j(y)$ for all $1\le j\le m$.
A solution is \emph{Pareto-optimal} if it is not strictly dominated by any other solution.
The set of objective values of Pareto-optimal solutions is called the \emph{Pareto front}.
The NSGA-III algorithm, originally proposed by Deb and Jain~\cite{deb2013evolutionary}, 
belongs to the evolutionary optimization framework and incorporates external mechanisms to support diversity maintenance, which can also help reduce the computational burden. 
Instead of exhaustively exploring the entire search space for Pareto-optimal solutions,  the algorithm initiates multiple predefined, targeted searches.
More specifically, NSGA-III initializes with a random population of size $N_{mo}$.
In each subsequent iteration, one generates an offspring population of size $N_{mo}$ using mutation and/or crossover operators. 
With a fixed population size, NSGA-III selects $N_{mo}$ individuals for the next iteration  from the combined pool of $2N_{mo}$ candidates. Since non-dominated solutions are used,  a ranking scheme is employed in which dominance serves as the primary criterion for survival. 
Individuals that are not strictly dominated by any other member of the population are assigned rank~1.
Subsequent ranks are determined recursively: each unranked individual that is strictly dominated only by those with ranks $1,\ldots,k-1$ is assigned rank $k$. 
Intuitively, the lower an individual's rank, the more interesting it is.
Denote $F_i$ be the set of individuals with rank $i$, and let $i^*$ be the smallest integer such that
\begin{align}
\sum_{i=1}^{i^*} |F_i| \ge N_{mo}.
\end{align}
All individuals with rank at most $i^* - 1$ are retained for the next generation. In addition, $0 < k \le N_{mo}$
individuals of rank $i^*$ must be selected so that the new population remains $N_{mo}$, allowing the next iteration to proceed.

\begin{algorithm}[H]
\caption{Nondominated Sorting Genetic Algorithm (NSGA-III)}
\label{alg:nsga3-pifdon}

\renewcommand{\algorithmicrequire}{\textbf{Require:}}
\renewcommand{\algorithmicensure}{\textbf{Ensure:}}

\begin{algorithmic}[1]
\Require Population size $N$, number of generations $G$, reference directions $R$
\Ensure Optimized Pi-FDON models on Pareto front

\State Initialize population $P_0$ with $N$ candidate Pi-FDON models
\State Evaluate objectives $\mathbf{F}(x)$ for all $x \in P_0$

\For{$t = 0$ \textbf{to} $G-1$}
    \State Generate offspring $Q_t$ using crossover and mutation operation
    \State Evaluate objectives $\mathbf{F}(x)$ for all $x \in Q_t$
    \State $R_t \gets P_t \cup Q_t$ \Comment{Combine parent and offspring}
    \State Perform non-dominated sorting on $R_t$ to obtain fronts $F_1, F_2, \dots$
    
    \State Initialize $P_{t+1} \gets \emptyset$, $i \gets 1$
    
    \While{$|P_{t+1}| + |F_i| \leq N$}
        \State $P_{t+1} \gets P_{t+1} \cup F_i$
        \State $i \gets i+1$
    \EndWhile

    \State Normalize objectives of $F_i$
    \State Associate each solution in $F_i$ with closest reference direction in $R$
    \State Compute niche count for each reference direction
    
    \While{$|P_{t+1}| < N$}
        \State Select reference direction with minimum niche count
        \State Choose one associated solution (random if tie) and add to $P_{t+1}$
    \EndWhile
\EndFor
\State \Return Final Pareto front $P_G$
\end{algorithmic}
\end{algorithm}
\raggedbottom

\subsubsection{Refined Nondominated Sorting Genetic Algorithm (RNSGA-III)}

\begin{figure}[H]
    \centering
    \includegraphics[width=.55\linewidth]{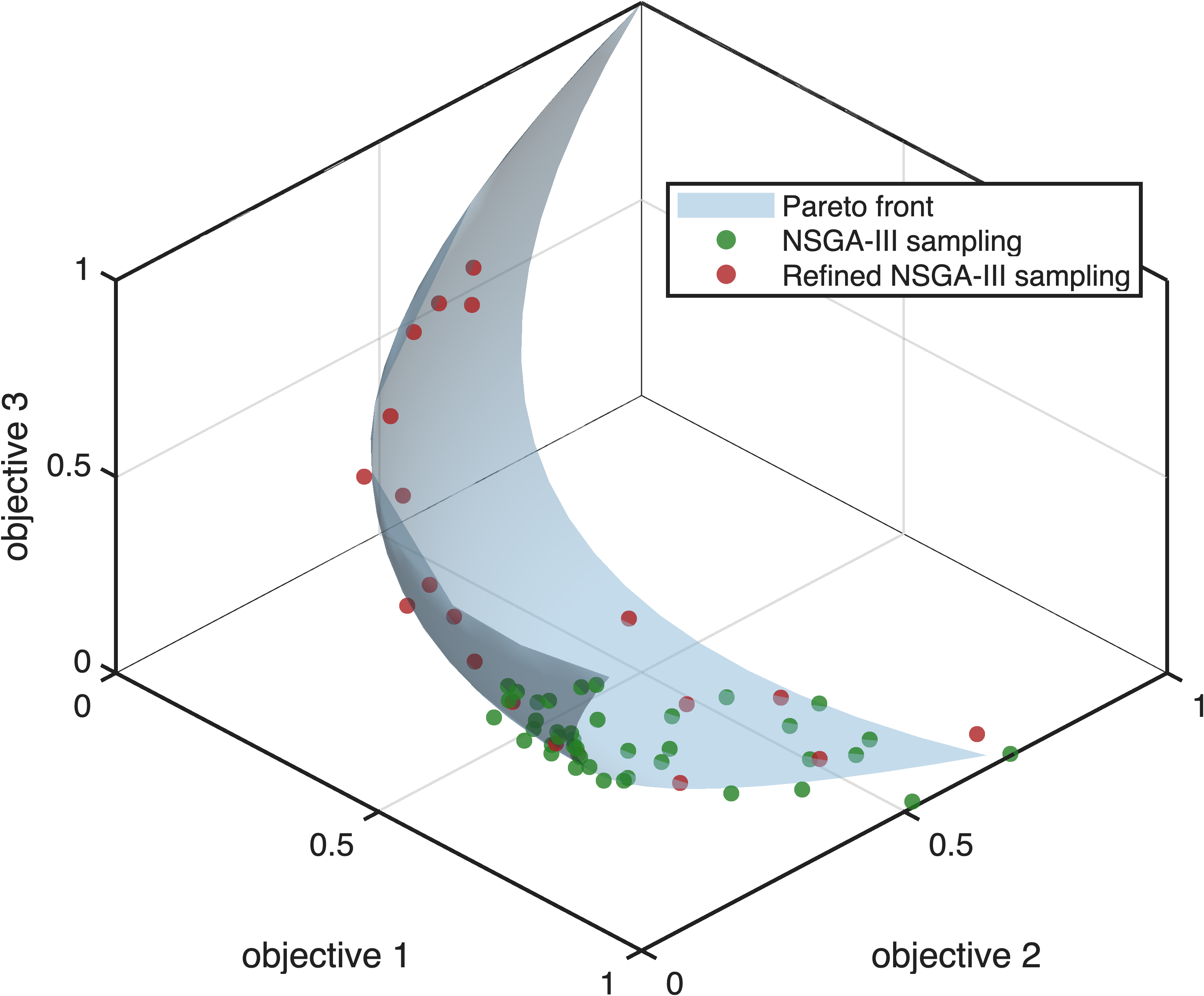}
    \caption{Illustration of NSGA-III and refined NSGA-III sampling near the Pareto front.
    \label{fig:nsga-points}}
\end{figure}

The vanilla NSGA-III algorithm (Algorithm~\ref{alg:nsga3-pifdon}), when applied to operator learning for PDE problems, aims to minimize the multi-objective loss in \eqref{loss-pi-don}. In principle, NSGA-III does not inherently ``contract'' samples to a local region of the front; in practice, however, it is common to observe that the non-dominated set concentrates on a local sub-arc (or sub-face) of the attainable front. This phenomenon is not a theoretical property of the algorithm, but typically arises from several factors: (i) a mismatch between the fixed reference directions and the shape/location of the optimal Pareto front, such that many directions do not intersect the feasible front \cite{hua2020generating,liu2019adaptation}; (ii) poor objective normalization or strong scale differences, which are common in operator-learning settings, introducing bias into the perpendicular-distance association step \cite{liang2019two}; (iii) high correlation among objectives (e.g., $\mathcal L_{\mathrm{PDE}}$ and $\mathcal L_{\mathrm{BC}}$) reducing the effective dimensionality of the front \cite{seada2015effect}; and (iv) feasibility/constraint handling overpowering diversity pressure, causing selection to repeatedly favor a small subset of niches \cite{yuan2014improved,farias2025non}. 
Stochastic mini-batching of residual and boundary losses can further amplify these effects because the resulting noise may affect both dominance checks (comparisons that determine whether one solution is better than another) and association steps (assigning solutions to reference directions or niches). Figure~\ref{fig:nsga-points} illustrates such a case, where vanilla NSGA-III produces sampling points (green) concentrated within a local sub-face of the attainable Pareto front.

To overcome this limitation, we propose refined NSGA-III, which integrates a refined Pareto sampling (RPS) strategy into the original NSGA-III framework. At each iteration, RPS first identifies solutions with sufficiently distinct objective values and then selects a representative subset based on pairwise distances between their outputs. The details of RPS are given in Algorithm~\ref{alg:pareto-thinning}. The red points in Figure~\ref{fig:nsga-points} show that RPS successfully disperses solutions across the Pareto surface, in contrast to the vanilla NSGA-III sampling (green points). 

\begin{algorithm}[H]
\caption{Refined Pareto Sampling (RPS)}
\label{alg:pareto-thinning}

\begin{algorithmic}[1]
\Require Pareto front $\mathcal{P} \in \mathbb{R}^{M \times K}$, number of models to select $N_{\text{sel}}$, optional distance threshold $\epsilon$
\Ensure Selected set of models $\mathcal{S}$

\State Initialize $\mathcal{S} \gets \emptyset$

\Comment{Step 1: Select extreme models}
\For{$k = 1$ \textbf{to} $K$}
    \State $i^* \gets \arg\min_{i} \mathcal{P}[i,k]$ \Comment{Model with lowest objective $k$}
    \State $\mathcal{S} \gets \mathcal{S} \cup \{\mathcal{P}[i^*]\}$
\EndFor

\State $\mathcal{R} \gets \mathcal{P} \setminus \mathcal{S}$ \Comment{Remaining models}

\Comment{Step 2: Select diverse models}
\While{$|\mathcal{S}| < N_{\text{sel}}$ \textbf{and} $\mathcal{R} \neq \emptyset$}
    \ForAll{model $r \in \mathcal{R}$}
        \State $d_r \gets \min_{s \in \mathcal{S}} \text{distance}(\mathcal{P}[r,:], s)$
    \EndFor
    
    \State $r^* \gets \arg\max_r d_r$ \Comment{Farthest model from selected set}
    
    \If{$d_{r^*} > \epsilon$ \textbf{or} $\epsilon$ not defined}
        \State $\mathcal{S} \gets \mathcal{S} \cup \{\mathcal{P}[r^*]\}$
    \EndIf
    \State $\mathcal{R} \gets \mathcal{R} \setminus \{\mathcal{P}[r^*]\}$
\EndWhile

\State \Return $\mathcal{S}$
\end{algorithmic}
\end{algorithm}

\subsection{ Replica Exchange Stochastic Gradient Langevin dynamics} \label{subsec:sgld-re}
Replica exchange stochastic gradient Langevin dynamics (reSGLD)~\cite{deng2020,deng2022adaptively,deng2020contour} extends the classical stochastic gradient Langevin dynamics (SGLD)~\cite{Welling11} by incorporating the replica exchange mechanism from statistical physics. While SGLD is a scalable Markov Chain Monte Carlo (MCMC) method that interpolates between stochastic optimization and posterior sampling as the step size decreases, its efficiency is often limited by the pathological curvature of the underlying energy landscape that may lead to local trapping. To reconcile this issue, reSGLD employs multiple diffusion processes at different temperatures and operates occasional swaps which can enable high-temperature replicas to traverse energy barriers and facilitate exploration across distinct local modes. This mechanism which combines the scalability of SGLD with the efficiency of replica exchange leads to accelerated convergence, which has been both theoretically established~\cite{Paul12, dong2020spectral} and empirically validated~\cite{deng2020}.

To illustrate the reSGLD algorithm, we first recall the negative log-posterior (energy function) given by
\begin{equation}
L(\boldsymbol{\beta})= -\log p(\boldsymbol{\beta}) - \sum_{i=1}^N \log P(\boldsymbol{d}_i|\boldsymbol{\beta}),
\end{equation}
where $p(\boldsymbol{\beta})$ is the prior and the second term corresponds to the complete-data log-likelihood. For large datasets, direct evaluation is expensive and sometimes prohibitive. A common remedy is to approximate it using a mini-batch $\mathcal{B}=\{\boldsymbol{d}_{s_i}\}_{i=1}^n$, which yields an unbiased stochastic energy estimator:
\begin{equation}
\widetilde L(\boldsymbol{\beta}) = -\log p(\boldsymbol{\beta})-\frac{N}{n}\sum_{i=1}^n \log P(\boldsymbol{d}_{s_i}|\boldsymbol{\beta}).
\end{equation}
The stochastic gradient Langevin dynamics (SGLD) iterates as
\begin{equation}
\boldsymbol{\beta}^{(k+1)}=\boldsymbol{\beta}^{(k)} - \eta^{(k)} \nabla \widetilde{L}(\boldsymbol{\beta}^{(k)})+\sqrt{2\eta^{(k)}\tau_1}\,\boldsymbol{\xi}^{(k)},
\end{equation}
where $\eta^{(k)}$ is the learning rate and $\boldsymbol{\xi}^{(k)}$ is a standard Gaussian noise. As $\eta^{(k)} \to 0$, SGLD converges to the invariant distribution $\pi(\boldsymbol{\beta})\propto \exp(-L(\boldsymbol{\beta})/\tau_1)$.
A naive extension of replica exchange Langevin dynamics (reLD) to the stochastic gradient setting leads to the following two-chain system:
\begin{equation}
\label{naive_resgld_eq}
\begin{split}
    \boldsymbol{\beta}^{(1,k+1)} &= \boldsymbol{\beta}^{(1,k)}- \eta^{(k)} \nabla \widetilde L(\boldsymbol{\beta}^{(1,k)})+\sqrt{2\eta^{(k)}\tau_1} \,\boldsymbol{\xi}^{(1,k)}, \\
    \boldsymbol{\beta}^{(2,k+1)} &= \boldsymbol{\beta}^{(2,k)}- \eta^{(k)} \nabla \widetilde L(\boldsymbol{\beta}^{(2,k)})+\sqrt{2\eta^{(k)}\tau_2} \,\boldsymbol{\xi}^{(2,k)},
\end{split}
\end{equation}
where $\tau_1 < \tau_2$ are different temperatures. Replica exchange is then attempted by swapping the two chains with a naive stochastic swapping rate
\begin{equation}
\label{swap_naive_eq}
\mathbb{S}(\boldsymbol{\beta}^{(1,k+1)}, \boldsymbol{\beta}^{(2,k+1)})
= \exp\!\left[\left(\frac{1}{\tau_1}-\frac{1}{\tau_2}\right)\big(\widetilde L(\boldsymbol{\beta}^{(1,k+1)})-\widetilde L(\boldsymbol{\beta}^{(2,k+1)})\big)\right].
\end{equation}
However, the stochastic estimates $\widetilde L(\cdot)$ enter \eqref{swap_naive_eq} through a nonlinear exponential, which introduces significant bias in the swap acceptance probability. This bias breaks detailed balance and results in poor approximation of the target distribution.
To address this, a correction is introduced by modeling the stochastic energy difference as a Gaussian perturbation,
\begin{equation}
\widetilde L(\boldsymbol{\beta}^{(1)})-\widetilde L(\boldsymbol{\beta}^{(2)})=
L(\boldsymbol{\beta}^{(1)})-L(\boldsymbol{\beta}^{(2)}) + \sqrt{2}\sigma W_1,
\end{equation}
where $W_1\sim \mathcal{N}(0,1)$. This yields the corrected swap acceptance rate
\begin{equation}
\label{sto_acceptance_eq}
\widetilde S_1=
\exp\!\left[\left(\frac{1}{\tau_1}-\frac{1}{\tau_2}\right)\Big(\widetilde L(\boldsymbol{\beta}^{(1)})-\widetilde L(\boldsymbol{\beta}^{(2)})
- \big(\tfrac{1}{\tau_1}-\tfrac{1}{\tau_2}\big)\sigma^2\Big)\right],
\end{equation}
which is now an unbiased estimator of the exact acceptance probability. The correction term $\left(\frac{1}{\tau_1}-\frac{1}{\tau_2}\right)\sigma^2$ compensates for the variance of the mini-batch noise and restores the validity of the replica exchange procedure.

\subsection{Uncertainty Quantification} \label{subsec:uq}
Uncertainty quantification (UQ) is critical to ensure the reliability of machine learning models in scientific applications. The reSGLD mechanism inherent in the Morephy-Net framework provides a natural and efficient way to perform Bayesian UQ. By sampling the posterior distribution of the model parameters $\theta$ using reSGLD, we obtain an ensemble of model predictions that can be used to quantify uncertainty. Specifically, the predictive mean and variance are given by $\bar{u}(x,t) = \frac{1}{N_s} \sum_{i=1}^{N_s} G_{\theta_i}(\kappa)(x,t) $ and variance $\sigma^2(x) = \frac{1}{N_s - 1} \sum_{i=1}^{N_s} (G_{\theta_i}(\kappa)(x,t) - \bar{u}(x,t))^2,
$ where $N_s$ denotes the sample size and $G_{\theta_i}(\kappa)(x,t)$ is the prediction from the $i$th sampled model.

\section{Numerical Results \label{sec:numeric}}
In this section, we will show the numerical test results for the proposed Morephy-Net in both forward and inverse problem setting.
Specifically, we consider two different test problems: the one-dimensional viscous Burgers' equation and the one-dimensional time-fractional mixed diffusion-wave equations (TFMDWEs). The inverse problem settings are formulated to identify the viscosity parameter in the Burgers’ equation and the fractional derivative order in the TFMDWEs. To avoid ambiguity, we provide a summary in Table \ref{tab:models_comparison} that lists all the models used for comparison with Morephy-Net in the numerical tests, along with their full names.

\begin{table*}[htbp]
\footnotesize
\centering
\caption{
Summary of models used in numerical tests.
Acronyms: DeepONet = \uline{Deep} \uline{O}perator \uline{Net}work;
PI-DON = \uline{P}hysics-\uline{i}nformed \uline{D}eep \uline{O}perator \uline{N}etwork;
PI-FDON = \uline{P}hysics-\uline{i}nformed \uline{F}ourier \uline{D}eep \uline{O}perator \uline{N}etwork;
Morephy-Net = \uline{M}ulti-\uline{O}bjective \uline{R}eplica \uline{E}xchange \uline{Phy}sics-informed Deep \uline{Ope}ra\uline{tor} \uline{Net}work.
}
\label{tab:models_comparison}
\renewcommand{\arraystretch}{1.2}
\begin{tabularx}{\textwidth}{lccc}
\hline
\textbf{Acronym} & \textbf{Physics-informed} & \textbf{Uncertainty Quantification} & \textbf{Fourier Feature}\\
\hline
DeepONet     & --   & --   & --   \\
PI-DON       & Yes  & --   & --   \\
PI-FDON      & Yes  & --   & Yes  \\
Morephy-Net  & Yes  & Yes  & Yes  \\
\hline
\end{tabularx}
\end{table*}

\subsection{Burgers' Equation}
The one-dimensional viscous Burgers' equation is a nonlinear partial differential equation frequently used as a benchmark \cite{lu2021deepxde,lu2021physics,raissi2019physics,mou2021data}. The equation with Dirichlet boundary conditions, is defined on the spatial domain $\Omega = [-1,1]$ and temporal domain $[0,T]$ given by the following:
\begin{align}
&\frac{\partial u}{\partial t} - \nu \frac{\partial^2 u}{\partial x^2} + u \frac{\partial u}{\partial x} = 0,\quad x \in \Omega,\; t \in [0,T],\label{eq:burgers}\\
&u(x,t) = 0,\quad \forall x \in \partial\Omega,\\
&u(x,0) = -\sin(\pi x).\vspace*{-.1cm}
\end{align}
Here, $u(x,t)$ represents the solution over space and time, and $\nu$ is the viscosity chosen to be $0.01/\pi$. 
\begin{remark}
           The Morephy-Net employs an ensemble strategy, which can be computationally expensive. This overhead, however, can be alleviated through parallelization, as each ensemble member is trained independently. The number of tasks/actors launched equals the population size. Furthermore, the incorporation of the evolutionary multi-objective optimization algorithm (NSGA-III) accelerates the convergence of each ensemble member relative to a standard DeepONet. As a result, while the computational cost of Morephy-Net is higher, it is not prohibitive. Table~\ref{tab:walltime_comparison}  list the wall time for training DeepONet and the proposed Morephy-Net. The wall time is recorded until the model reaches a convergence plateau. All experiments were performed on a linux machine with an NVIDIA H100 GPU (80G). 

        \begin{table}[htbp]
        \footnotesize
\centering
\caption{Comparison of model performance for Wall time of training and $L^2$ relative error (RE) between DeepONet and Morephy-Net.
}
\label{tab:walltime_comparison}
\begin{tabular}{lcc}
\toprule
\textbf{Method} & \textbf{Wall Time (s)} & \textbf{$L^2$ RE} \\
\midrule
DeepONet    & 270  & 0.0550 \\
\textbf{Morephy-Net} & \textbf{450 }& \textbf{0.0436} \\
\bottomrule
\end{tabular}
\end{table}
\end{remark}

\subsubsection{Hyperparameters}
To ensure a rigorous and unbiased comparison between DeepONet, PI-DON, and PI-FDON, we standardized the core architecture and training protocol across all models. For each network, both the branch and trunk sub-networks were constructed as Multi-Layer Perceptrons (MLPs) featuring a depth of 8 hidden layers and a width of 50 neurons per layer. The training regimen was also held constant for all experiments. We employed the Adam optimizer with a learning rate of 5e-4. To mitigate overfitting and improve training efficiency, an early stopping criterion was implemented, which terminated the training process if the validation loss did not demonstrate significant improvement over a predefined number of epochs. 

\subsubsection{Forward Problem}
\begin{figure*}[!t]
    \centering
 \subfloat[Spatiotemporal fields  \label{Burgers_3d_forward}]{\includegraphics[width=\linewidth]{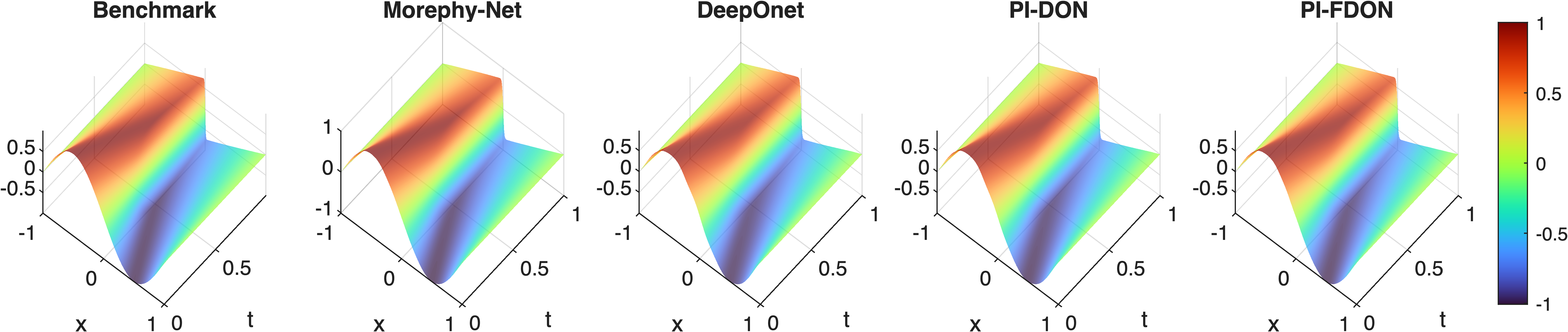}}
 \\
 \subfloat[Solution $u(x,t)$ at different fixed time $t= t_f$. The blue shadow shows the 95\% confidence Interval.  \label{Burgers_forward_UQ_result}
]{
 \includegraphics[width=.30\linewidth]{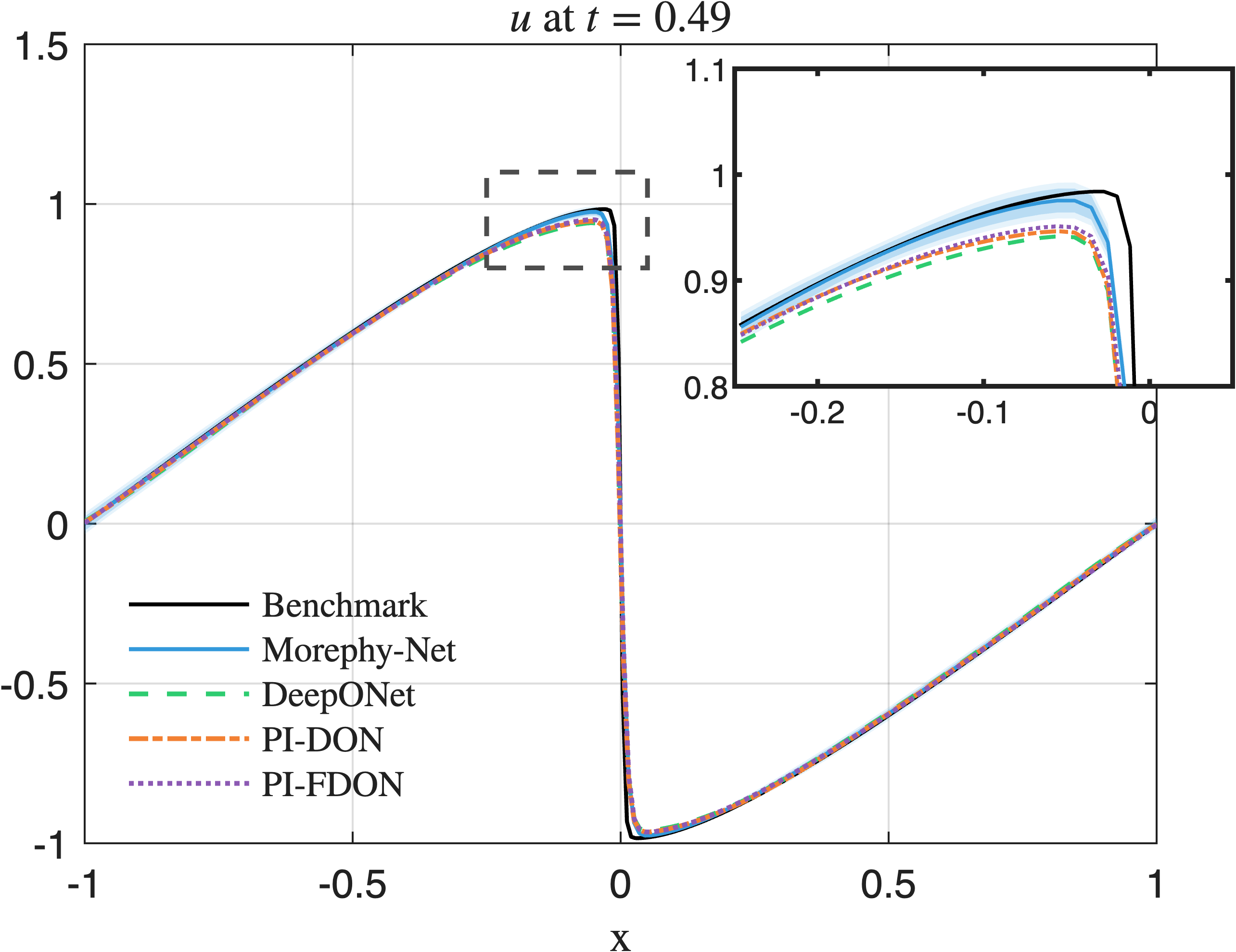}
 \hspace{.2cm}
 \includegraphics[width=.30\linewidth]{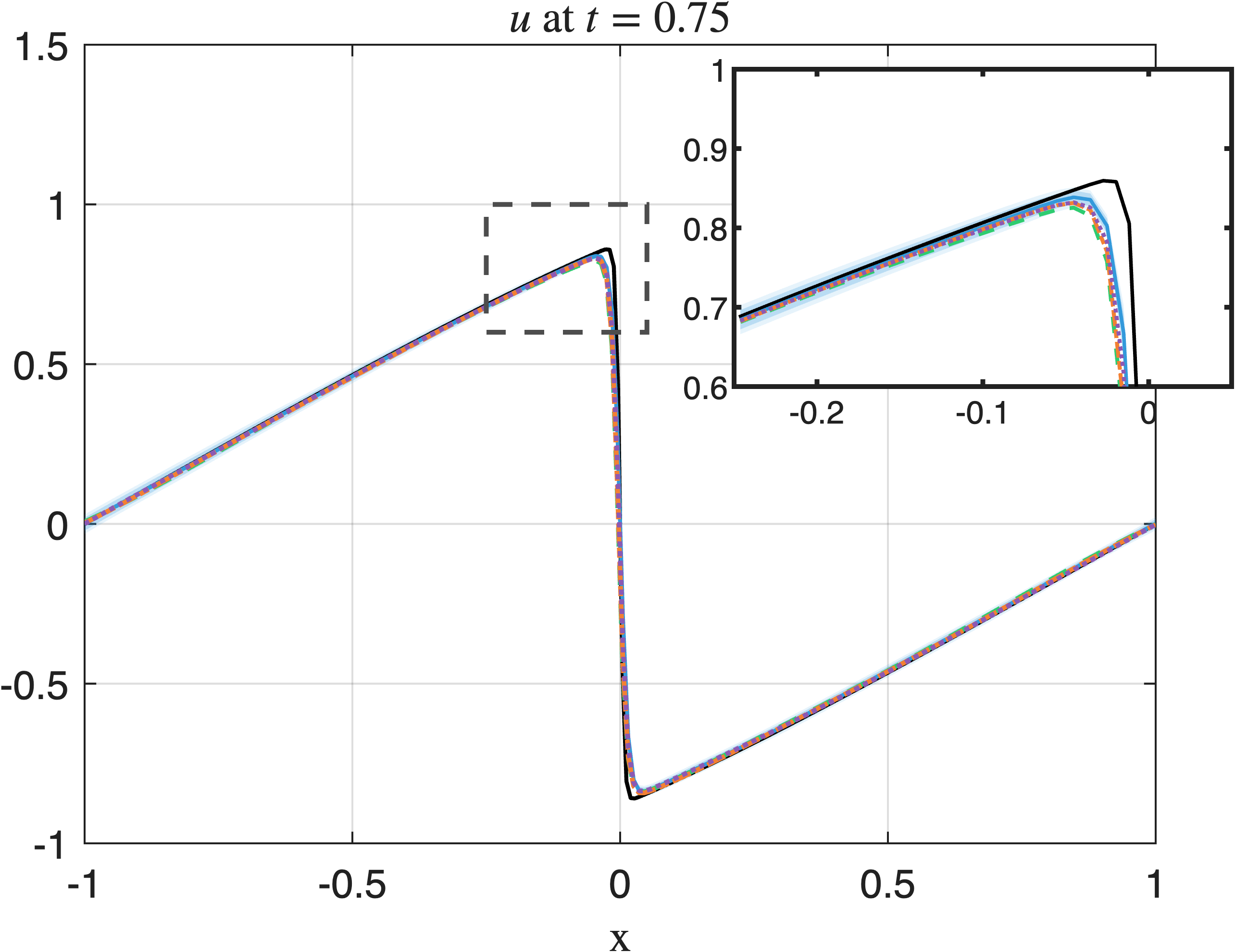}
 \hspace{.2cm}
 \includegraphics[width=.30\linewidth]{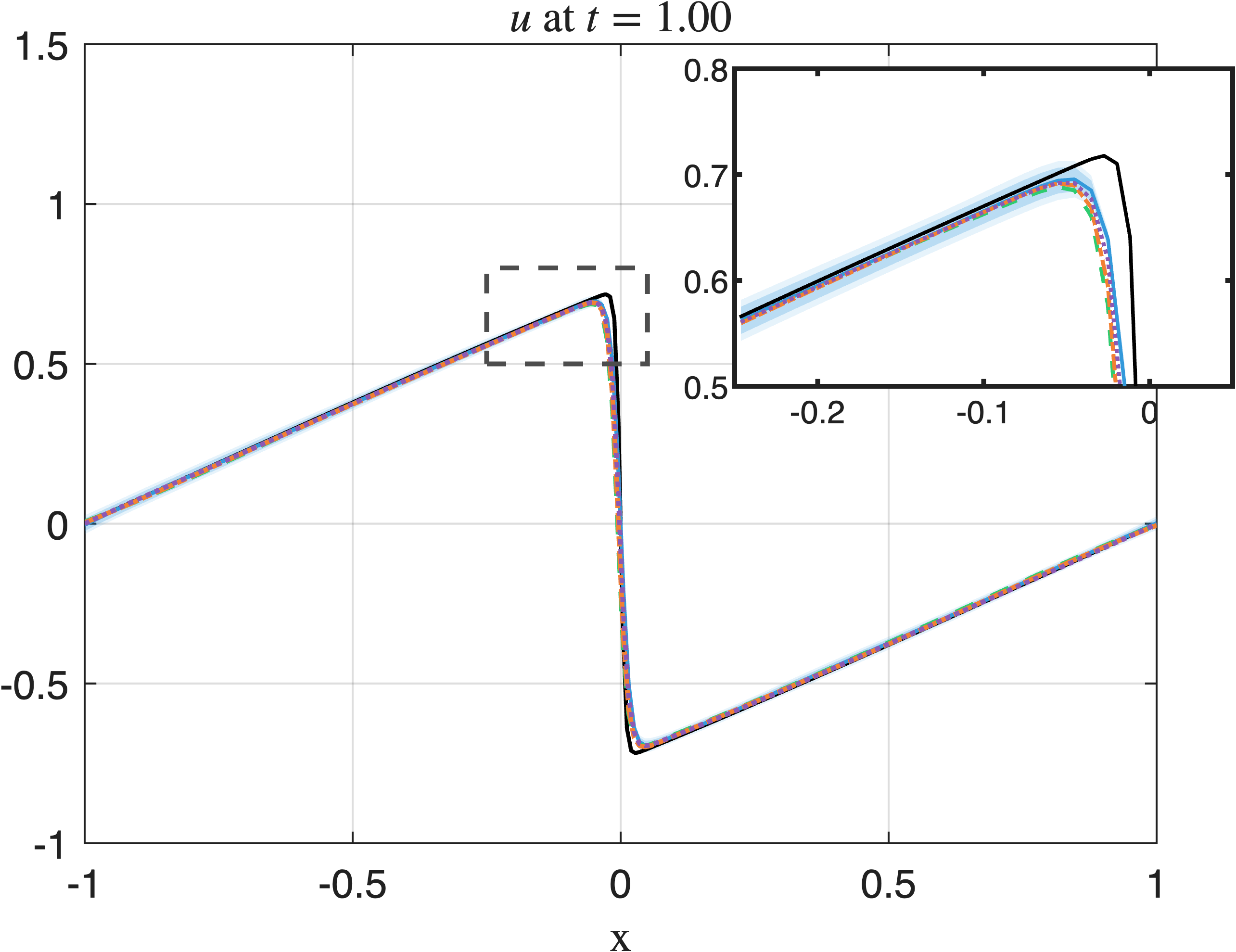}
 }
\\
 \subfloat[$L^1$ Errors     \label{Burgers_forward_result_error}
]{\includegraphics[width=\linewidth]{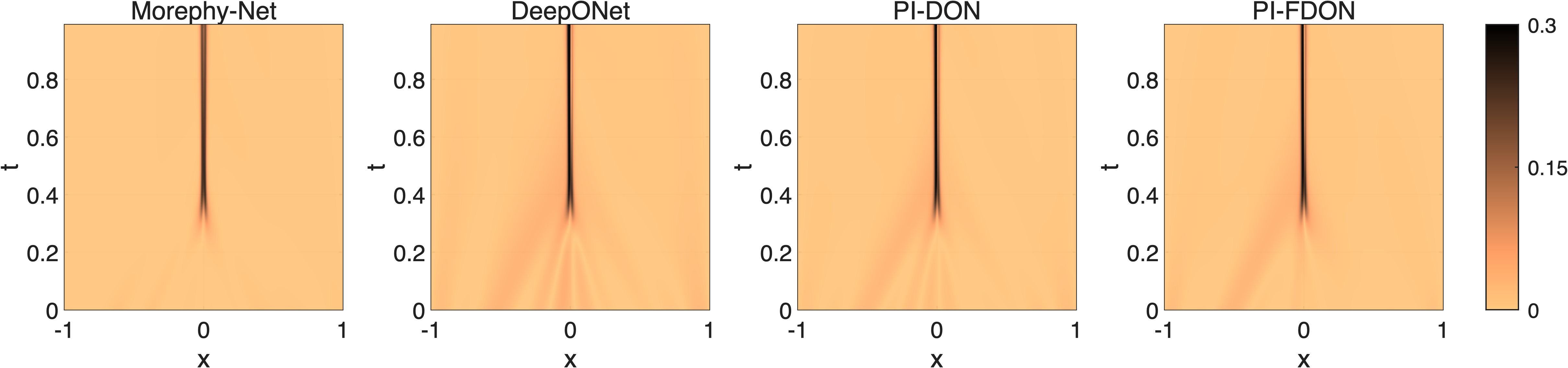}}

    \caption{Comparisons of different models (DeepONet, PI-DON, PI-FDON, and Morephy-Net) and benchmark solutions for the forward Burgers problem. Panel (a). Spatiotemporal fields of the solutions. Panel (b). Solutions at different fixed time instances $t=t_f$. Panel (c). $L^1$ errors.}
    \label{Burgers_forward_result_prediction}
\end{figure*}
We solve the forward problem for the 1D viscous Burgers' equation to evaluate the performance of four models: DeepONet, PI-DON, and PI-FDON. The goal is to predict the solution $u(x,t)$ for the entire spatio-temporal domain $(x,t)\in [-1,1] \times [0,1]$, given the initial and boundary conditions. The visualization results are presented in Figure~\ref{Burgers_forward_result_prediction}. As shown in first row~\ref{Burgers_3d_forward}, all four models successfully learn the overall dynamics of the Burgers' equation. The 3D spatiotemporal plots demonstrate that each model captures the smooth initial state and the subsequent formation of a steep shock front as time progresses. A more detailed and analysis is presented in second row~\ref{Burgers_forward_UQ_result} presents solution snapshots at time $t = 0.49, t =0.75 $ and $ t = 1.00$, revealing significant performance differences in resolving the shock front discontinuity. The prediction from Morephy-Net, shown as a blue shaded region representing the $95\%$ confidence interval, almost perfectly overlaps the benchmark solution. It captures the sharp shock wave with no visible overshooting or spurious oscillations, yielding a stable and physically plausible result. In the contrast, the other three models struggle to resolve the sharp gradient. As highlighted in the insets, these models exhibit  non-physical oscillations near the shock, an issue that becomes more pronounced as the shock steepens over time. The $L^1$ error distributions are plotted in third row~\ref{Burgers_forward_result_error}. The errors for DeepOnet, PI-DON, and PI-FDON are heavily concentrated along the trajectory of the shock wave. The error map for Morephy-Net is visibly lighter, indicating a lower error magnitude across the domain and reinforcing its superior performance in shock capturing. The quantitative performance of the models is summarized in Table~\ref{tab:burgers_forward_model_comparison} to clearly illustrate the model difference. As from our observation, the proposed PI-FDON shows higher accuracy than DON and PI-DON model. Crucially, the Morephy-Net model demonstrates the best performance overall. It better satisfies the residual constraint and boundary constraint and also achieves the lowest $L^2$ relative error.

\begin{table}[htbp]
\footnotesize
\centering
\caption{Comparison of model performance for the Losses and $L^2$ relative error (RE) in the 1D Burgers Equation forward problem.}
\label{tab:burgers_forward_model_comparison}
\begin{tabular}{l c c c }
\toprule
\textbf{Model} & \textbf{Residual Loss} & \textbf{IBC Loss} & \textbf{$L^2$ RE}\\
\midrule
DeepONet      & N/A       & 0.0008      & 0.0550 \\
PI-DON   & 0.0007   & 0.0003     & 0.0510 \\
PI-FDON  & 0.0004   & 0.0003     & 0.0446 \\
\textbf{Morephy-Net}   & \textbf{0.0002}   & \textbf{0.0001}     & \textbf{0.0436} \\
\bottomrule
\end{tabular}
\end{table}

\subsubsection{Inverse Problem}
We now focus on the inverse problem for Burgers' equation, a more challenging task where the viscosity parameter, $v$, is treated as an unknown variable. Our objectives are to infer the correct value of the viscosity parameter $v$ and accurately reconstruct the full continuous solution field $u(x,t)$ over the entire spatial-temporal domain. This type of parameter inference poses a significant challenge for deep learning models. The complex and  non-convex loss landscapes associated with inverse problems mean that the optimization process can easily become trapped in poor local minima, leading to inaccurate parameter estimates and solution reconstructions. We use the observational data serve as anchors during the training process, ensuring that the network remains consistent with the underlying physical dynamics. In the inverse problem, the total loss function is composed of three components: the PDE residual loss, the boundary and initial condition loss, and the data loss.

\begin{figure*}[!t]
    \centering
 \subfloat[Spatiotemporal fields\label{Burgers_3d_inverse}]{\includegraphics[width=\linewidth]{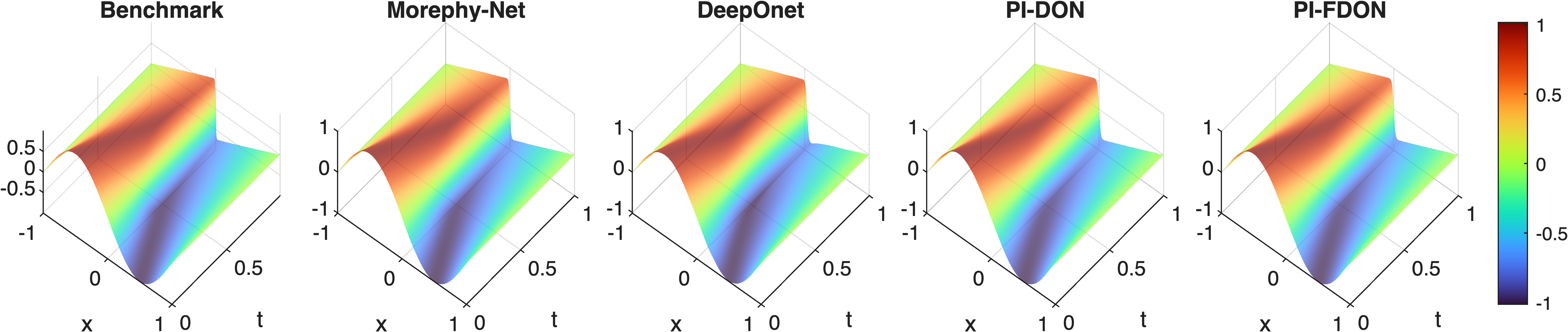}}
 \\
 \subfloat[Solution $u(x,t)$ at different fixed time $t= t_f$. The blue shading indicates the 95\% confidence interval.     \label{Burgers_inverse_UQ_result}
]{
 \includegraphics[width=.30\linewidth]{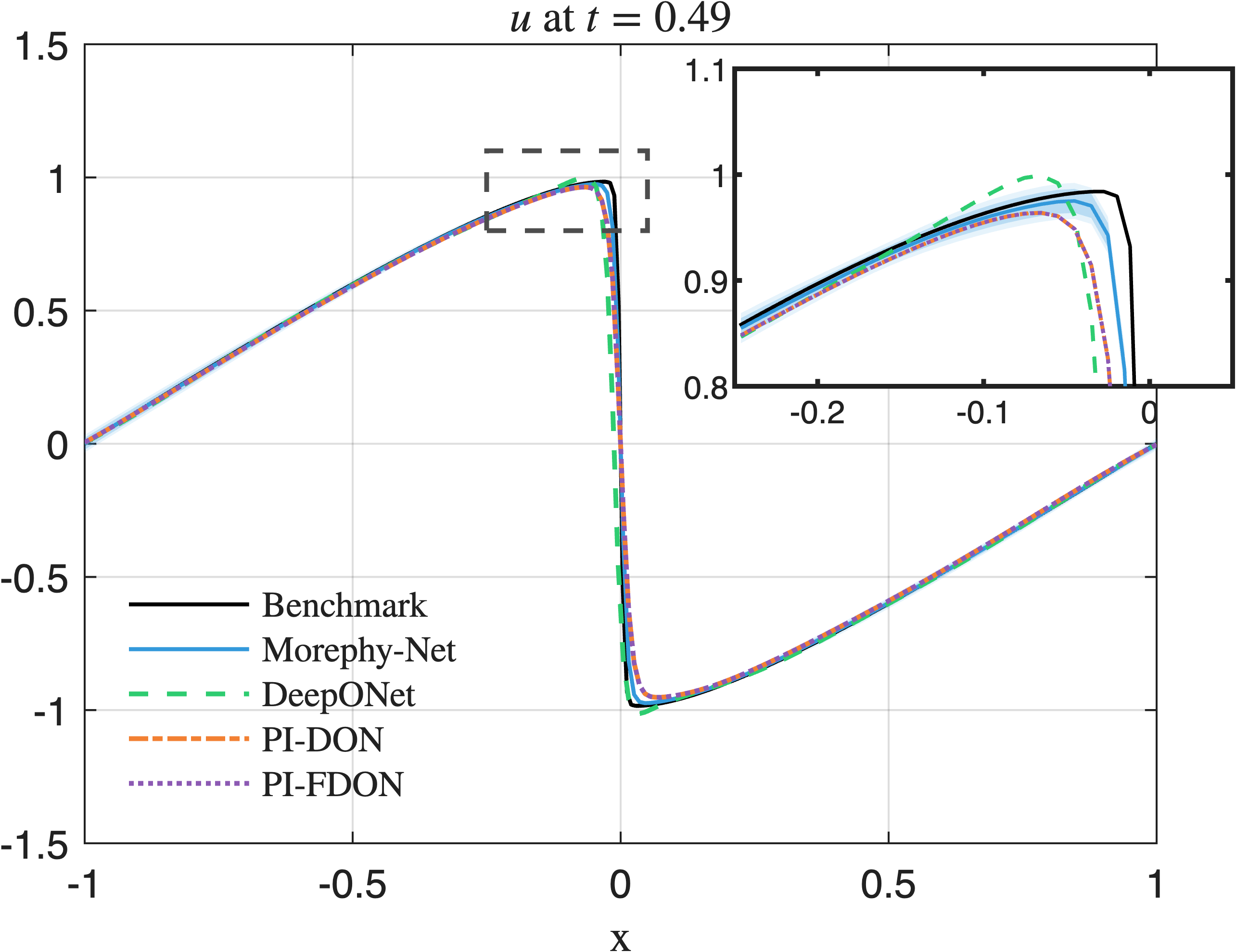}
 \hspace{.2cm}
 \includegraphics[width=.30\linewidth]{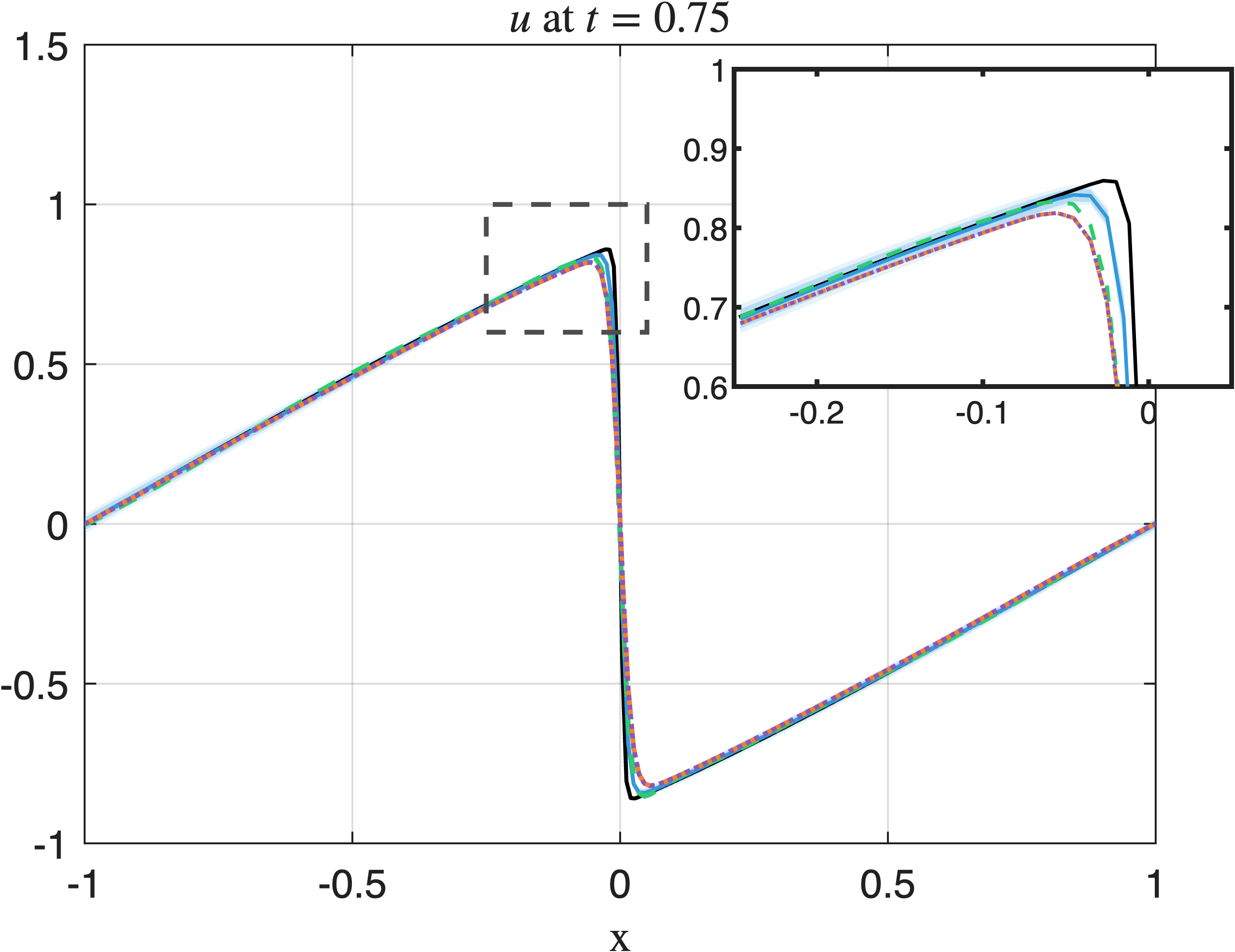}
 \hspace{.2cm}
 \includegraphics[width=.30\linewidth]{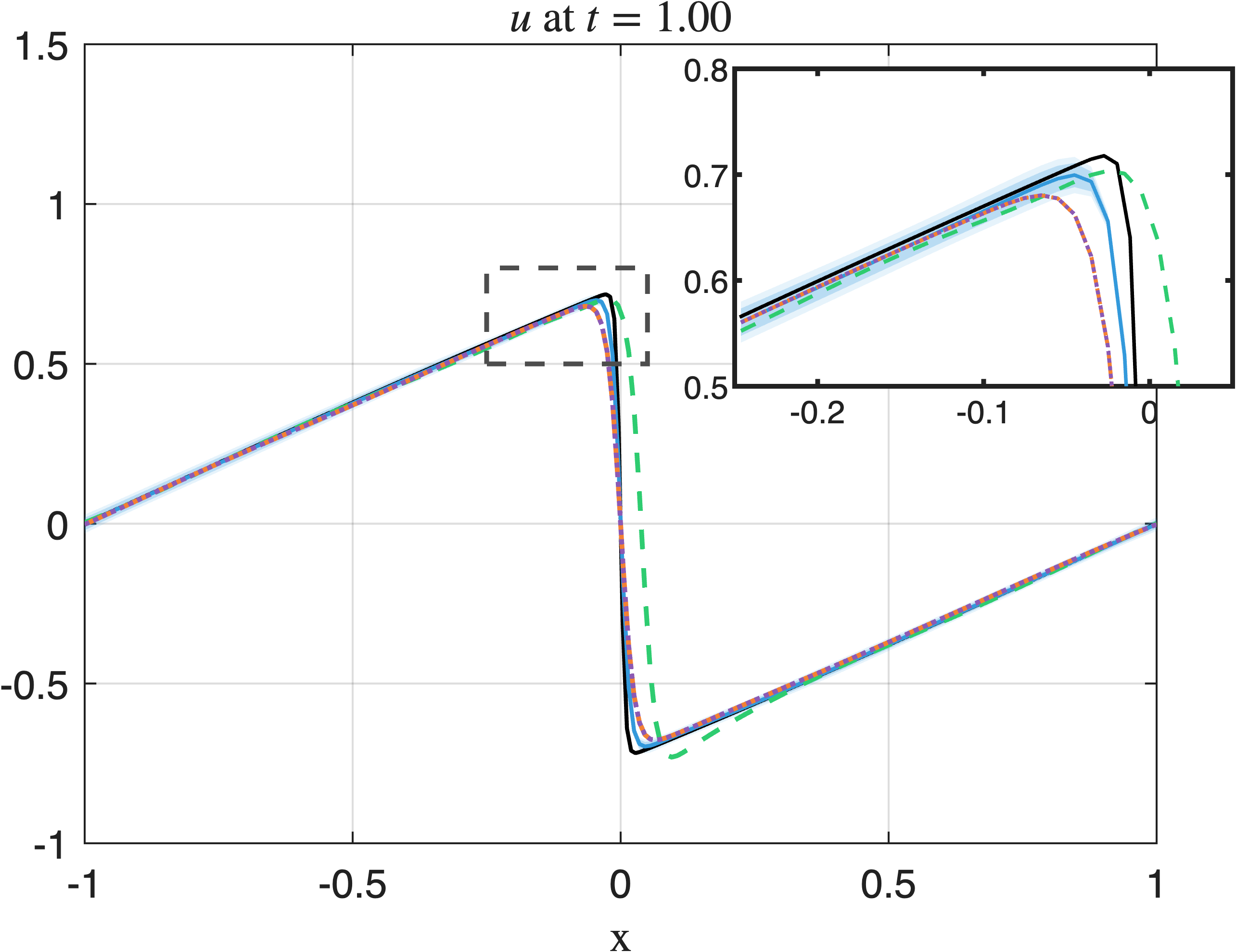}
 }
\\
 \subfloat[$L^1$ Errors     \label{Burgers_inverse_result_error}
]{\includegraphics[width=\linewidth]{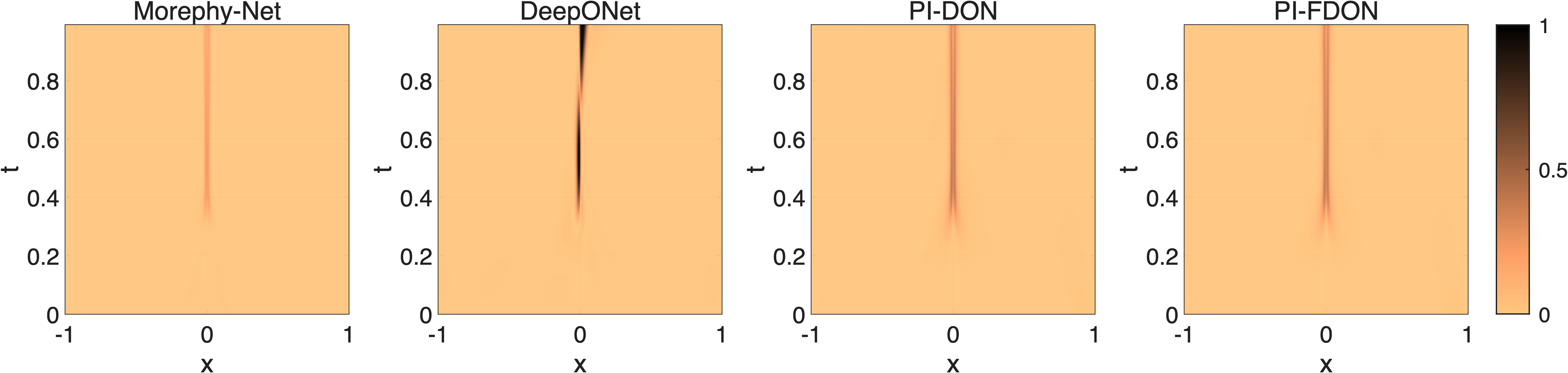}}

    \caption{Comparisons of different models (DeepONet, PI-DON, PI-FDON, and Morephy-Net) and benchmark solutions for the inverse Burgers problem. Panel (a). Spatiotemporal fields of the solutions. Panel (b). Solutions at different fixed time instances $t=t_f$.  Panel (c). $L^1$ errors. }
    \label{Burgers_inverse_result_prediction}
\end{figure*}

In Figure~\ref{Burgers_inverse_result_prediction}, the first row~\ref{Burgers_3d_inverse}, shows the reconstruction result in the spatial-temporal solution field. The second row~\ref{Burgers_inverse_UQ_result} shows the snapshot of the result at three distinct timesteps. As shown in this figure, the prediction of DeepONet shows the most severe oscillations, above the benchmark at $t = 0.49$ and $t = 1.00$ but below the benchmark at $t = 0.75$. PI-DON and Pi-FDON also fail to produce a clean, stable prediction. However, Morephy-Net demonstrates best performance in this scenario. The blue shaded region shows the $95\%$ confidence interval and it almost perfectly overlaps the black benchmark line. It has a great improvement when compare with the other three models. The third row~\ref{Burgers_inverse_result_error} displays a comparison of error heatmaps for four models. Each plot visualizes the absolute error between the model's prediction and the true solution over a 2D domain of space $x$ and time $t$. As we can observe from the figure that the Morephy-Net performs the best accuracy. the lighter bright line along the center indicates a lower error that is concentrated where the shock wave occurs.

\begin{table*}[htbp]
\footnotesize
\centering
\caption{Comparison of model performance for the Losses and $L^2$ relative error (RE) in the Burgers' inverse problem. For reference, the truth value of $v = 0.01/pi \approx 0.00318$.}
\label{tab:Burgers_Inverse_Problem_model}
\begin{tabular}{l c c c c c c}
\toprule
\textbf{Model} & \textbf{Residual} & \textbf{IBC loss} & \textbf{Data loss} & \textbf{Predicted $\boldsymbol{v}$} & \textbf{$L^2$ RE}\\
\midrule
DeepONet     & N/A &0.0007 & 0.0009& 0.0161     & 0.1174  \\
PI-DON   & 0.0010&0.0043 &0.0015 & 0.0098    & 0.0700 \\
PI-FDON  & 0.0008 & 0.0026 & 0.0008 &0.0073    & 0.0560 \\
 \textbf{Morephy-Net} & \textbf{0.0003} & \textbf{0.0005} &\textbf{0.0001}& \textbf{0.0062} & \textbf{0.0358} \\
\bottomrule
\end{tabular}
\end{table*}

\subsubsection{Inverse Problem With Noisy Data}
We now investigate a more challenging version of the inverse problem to assess model robustness. The sparse training data are corrupted by adding noise drawn from a Gaussian distribution with a zero mean. We introduce a substantial noise level by setting the noise's standard deviation to 50\% of the standard deviation of the true data, evaluating the models' capacity to learn the governing physics from corrupted observations. 

\begin{figure*}[!t]
    \centering
 \subfloat[Spatiotemporal fields]{\includegraphics[width=\linewidth]{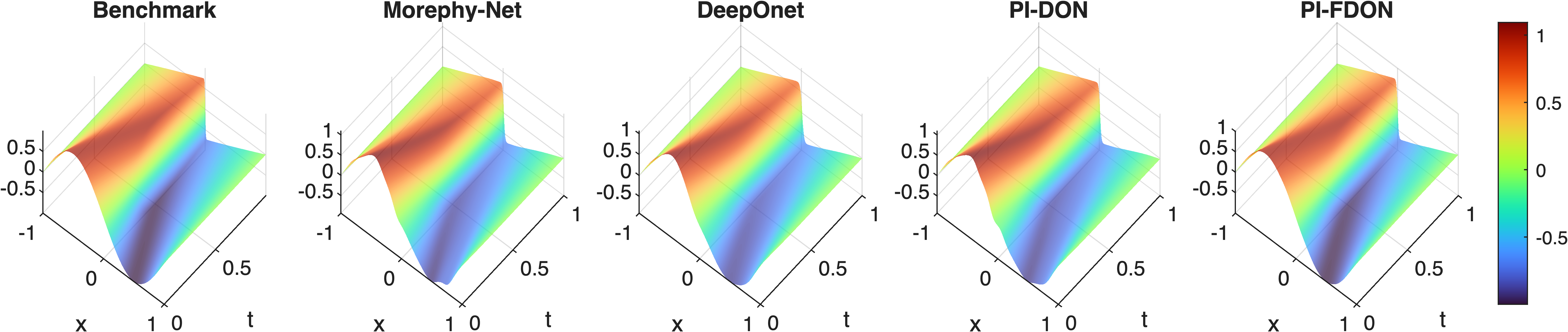}}
 \\
 \subfloat[Solution $u(x,t)$ at different fixed time $t= t_f$. The blue shading indicates the 95\% confidence interval.     \label{Burgers_inverse_noisy_UQ_result}
]{
 \includegraphics[width=.30\linewidth]{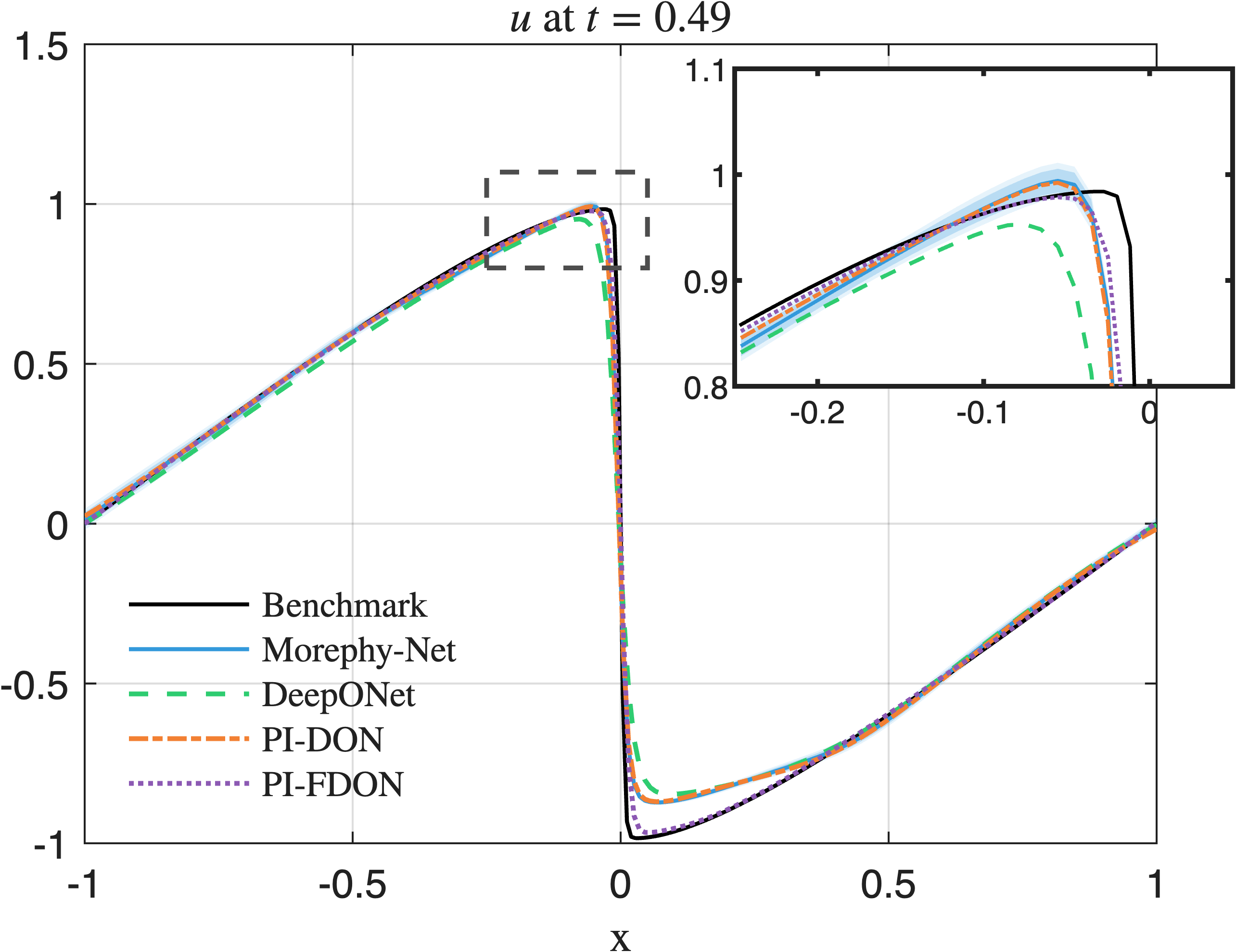}
 \hspace{.2cm}
 \includegraphics[width=.30\linewidth]{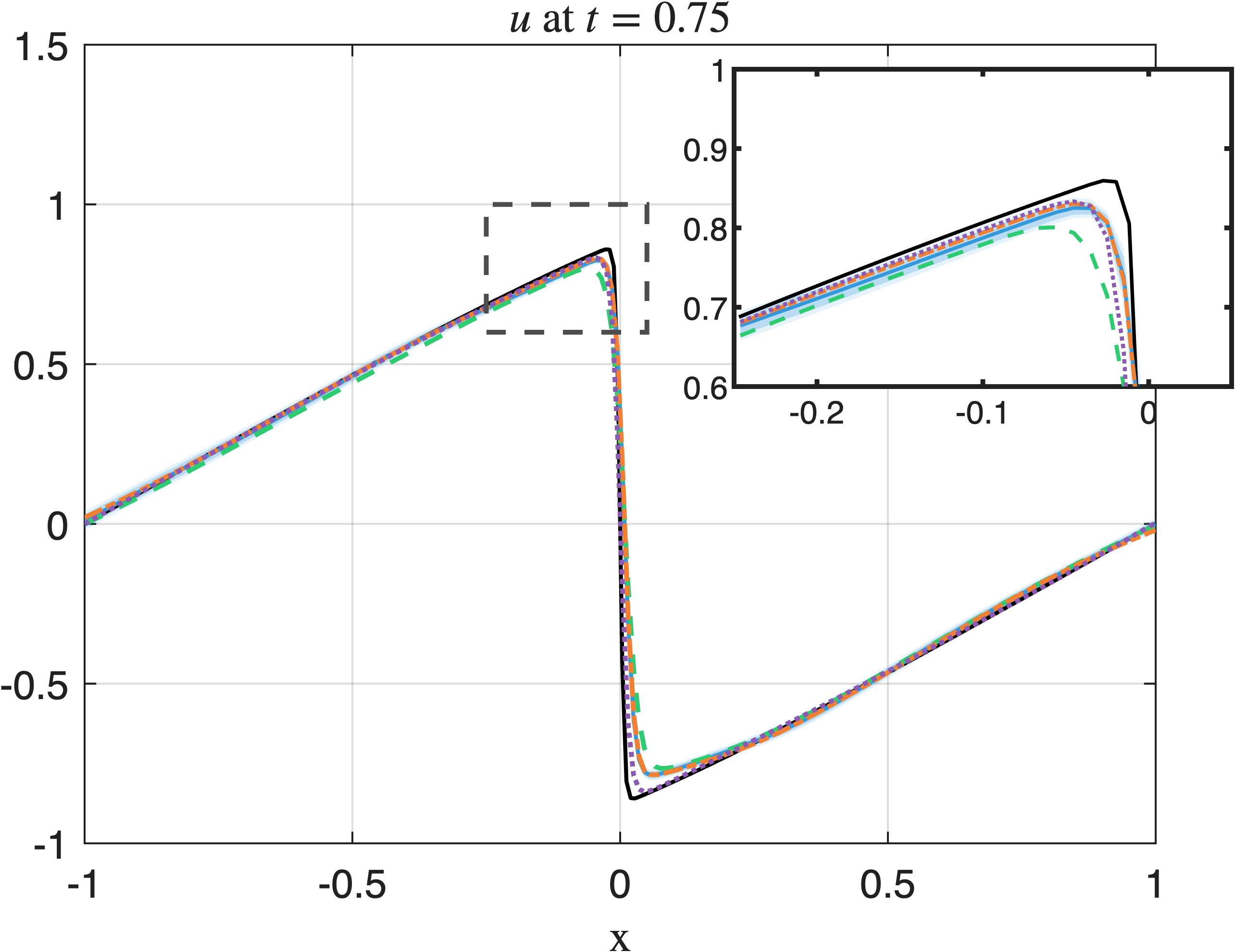}
 \hspace{.2cm}
 \includegraphics[width=.30\linewidth]{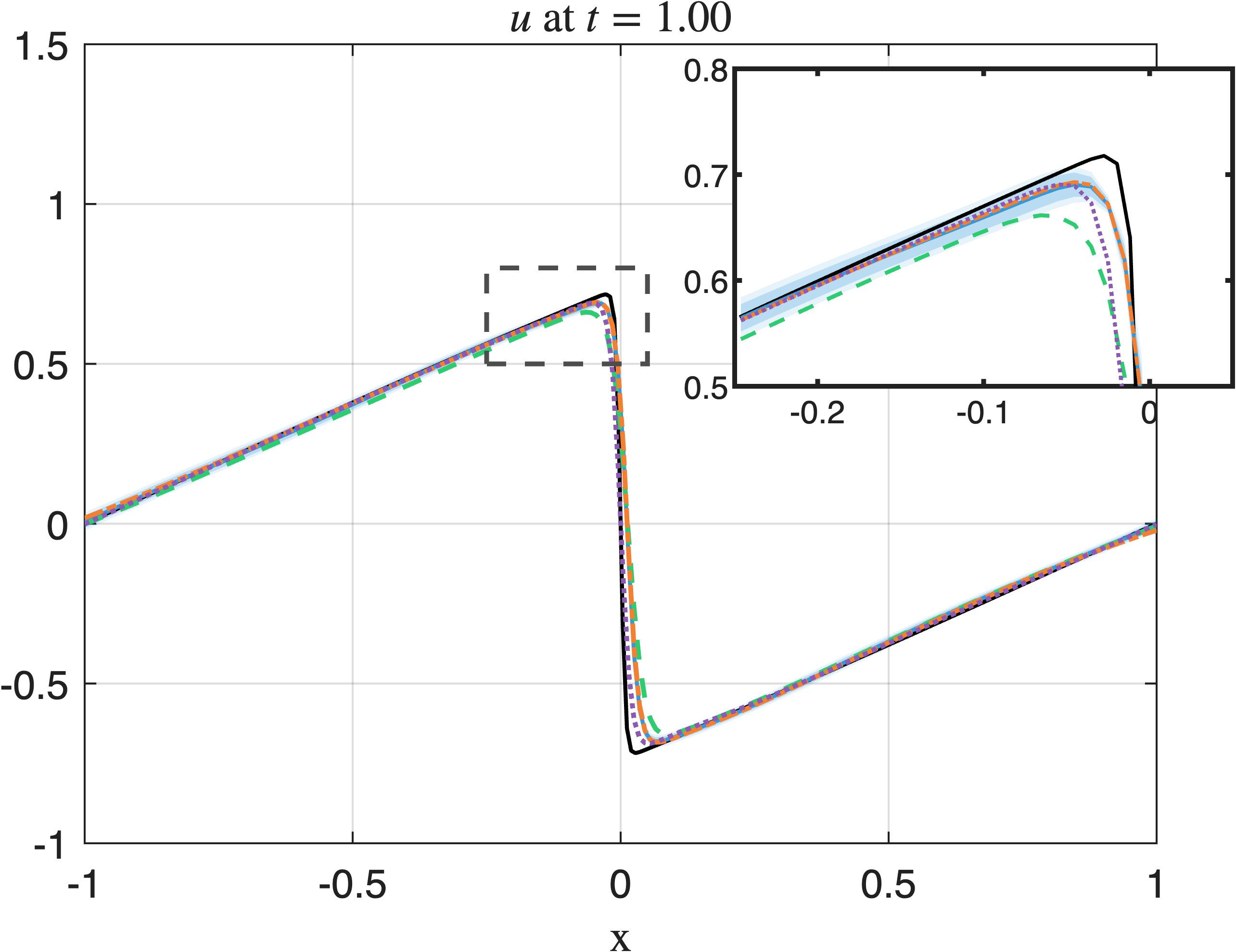}
 }
\\
 \subfloat[$L^1$ Errors     \label{Burgers_inverse_noisy_result_error}
]{\includegraphics[width=\linewidth]{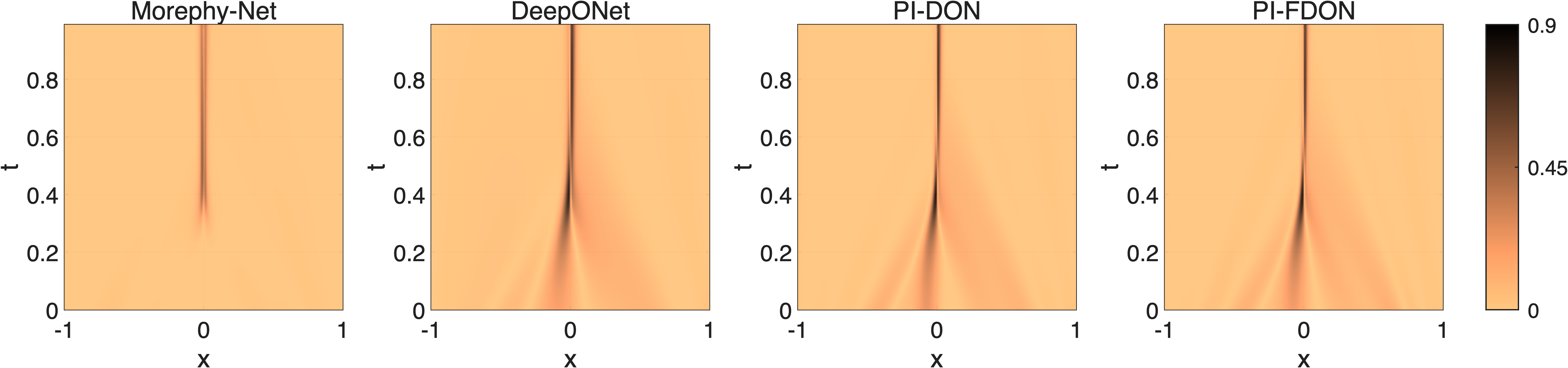}}

    \caption{Comparisons of different models (DeepONet, PI-DON, PI-FDON, and Morephy-Net) and benchmark solutions for the inverse Burgers problem with noisy observation data. Panel (a). Spatiotemporal fields of the solutions. Panel (b). Solutions at different fixed time instances $t=t_f$.  Panel (c). $L^1$ errors. }
    \label{Burgers_inverse_noisy_result_prediction}
\end{figure*}

Figure~\ref{Burgers_inverse_noisy_result_prediction} shows the reconstructed solution and error in the spatial-temporal domain. In the second row~\ref{Burgers_inverse_noisy_UQ_result}, the Morephy-Net model shows good performance and reliability. The prediction of Morephy-Net is visualized as a blue shaded area, representing the 95\% confidence interval in second row~\ref{Burgers_inverse_noisy_UQ_result}. As clearly shown in the zoomed in figure, this confidence interval covers black benchmark line from $x = -1$ to 0. In the contrast, the other three models fail to accurately capture the shock. Their point predictions exhibit significant deviations and spurious oscillations near the discontinuity. The spatial-temporal $L^1$ error distribution for four models are shown in the third row~\ref{Burgers_inverse_noisy_result_error}. The Morephy-Net error map is lighter than others, indicates a substantially lower prediction error across the entire space-time domain. Table~\ref{tab:Burgers_Inverse_Noise_Problem_model} summarizes the results for the challenging inverse problem with noisy data, where Morephy-Net shows the most robust performance. It not only achieves the best fit to the physical and data constraints but also yields the most accurate parameter prediction ($0.0063$). Consequently, it achieves a significantly lower $L^2$ relative error of 0.0744 compare with Pi-FDON of 0.1249.

\begin{table*}[htbp]
\footnotesize
\centering
\caption{Comparison of model performance for the Losses and $L^2$ relative error (RE) in the Burgers' inverse problem with noisy data.}
\label{tab:Burgers_Inverse_Noise_Problem_model}
\begin{tabular}{l c c c c c c }
\toprule
\textbf{Model} & \textbf{Residual Loss} & \textbf{IBC loss} & \textbf{Data loss} & \textbf{Predicted $\boldsymbol{v}$} & \textbf{$L^2$ RE}\\
\midrule
DeepONet     & N/A &0.0068 &0.5762& 0.0086     & 0.1421  \\
PI-DON   & 0.0084&0.0042 &0.3461& 0.0078   & 0.1260 \\
PI-FDON  & 0.0074  & 0.0026& 0.4653 & 0.0073    & 0.1249 \\
\textbf{Morephy-Net}  & \textbf{0.0043}       & \textbf{0.0021}&\textbf{0.2713}    & \textbf{0.0063}       & \textbf{0.0744} \\
\bottomrule
\end{tabular}
\end{table*}

\subsection{Time-fractional Mixed Diffusion-Wave Equations (TFMDWEs)}
In this section, we consider the \textit{time-fractional mixed diffusion-wave equations} (TFMDWEs), which generalize classical diffusion and wave equations by incorporating fractional-order time derivatives \cite{liu2019alternating,du2021temporal,lu5080263fpinn}. The TFMDWEs with Dirichlet boundary conditions are defined as follows:
\begin{align}
& D_t^{\alpha} u(x,t) = \frac{\partial^2 u}{\partial x^2} + f(x,t), \quad t \in [0,1],\; x \in [0,\pi],\\
&u(x,t) = 0,\quad \forall x \in \partial\Omega,\\
& u(x, 0) = 0, \quad x \in \Omega,
\end{align}
where the fractional order \(\alpha\) yields:
$
\alpha \in [0,1], \quad \forall t \in [0,1],
$ and the forcing term \(f(x,t)\) is explicitly defined as:
\begin{align}
f(x,t) = \frac{\Gamma(4)}{\Gamma(4-\alpha)}\, t^{3-\alpha}\sin(x) + t^3\sin(x),
\end{align}
with $\Gamma(\cdot)$ denoting the Gamma function. This equation introduces fractional-order temporal dynamics, combining features of both diffusion and wave phenomena. The fractional order $\alpha$ controls the transition between diffusive and wave-like behaviors, making TFMDWEs particularly useful in modeling complex systems exhibiting anomalous transport and non-local temporal interactions.

\subsubsection{Hyperparameters}
To ensure a fair comparison between DeepONet, PI-DON, and PI-FDON, we standardized the core architecture and training protocol across all models. For each network, both the branch and trunk sub-networks were constructed as Multi-Layer Perceptrons (MLPs) featuring a depth of 3 hidden layers and a width of 100 neurons per layer. The training regimen was also held constant for all experiments. We employed the Adam optimizer with a learning rate of 5e-4. The early stopping criterion was implemented to mitigate overfitting and improve training efficiency.

\subsubsection{Forward Problem}
\begin{figure*}[!t]
    \centering
 \subfloat[Spatiotemporal fields\label{FPDE_forward_result}]{\includegraphics[width=\linewidth]{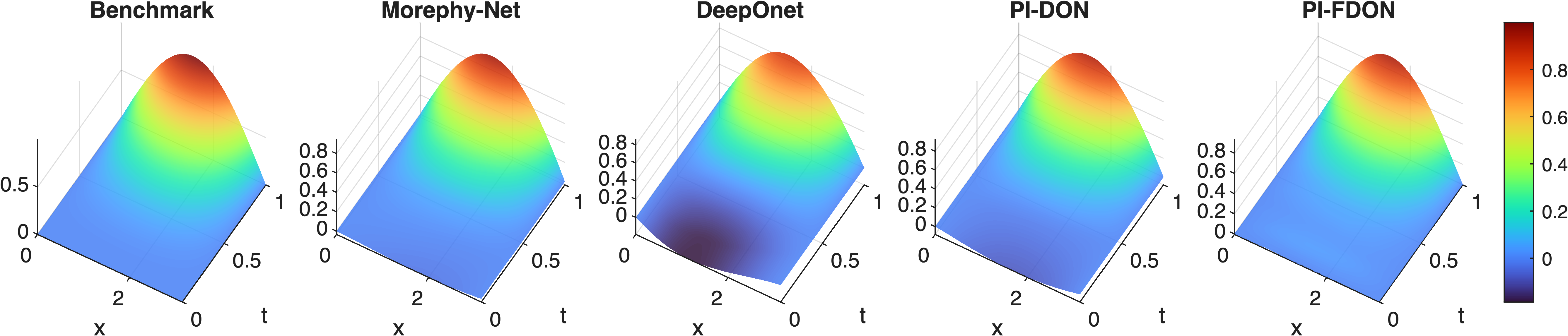}}
 \\
 \subfloat[Solution $u(x,t)$ at different fixed location $x= x_f$. The blue shading indicates the 95\% confidence interval.     \label{FPDE_forward_UQ_x_result}
]{
 \includegraphics[width=.18\linewidth,height=.16\linewidth]{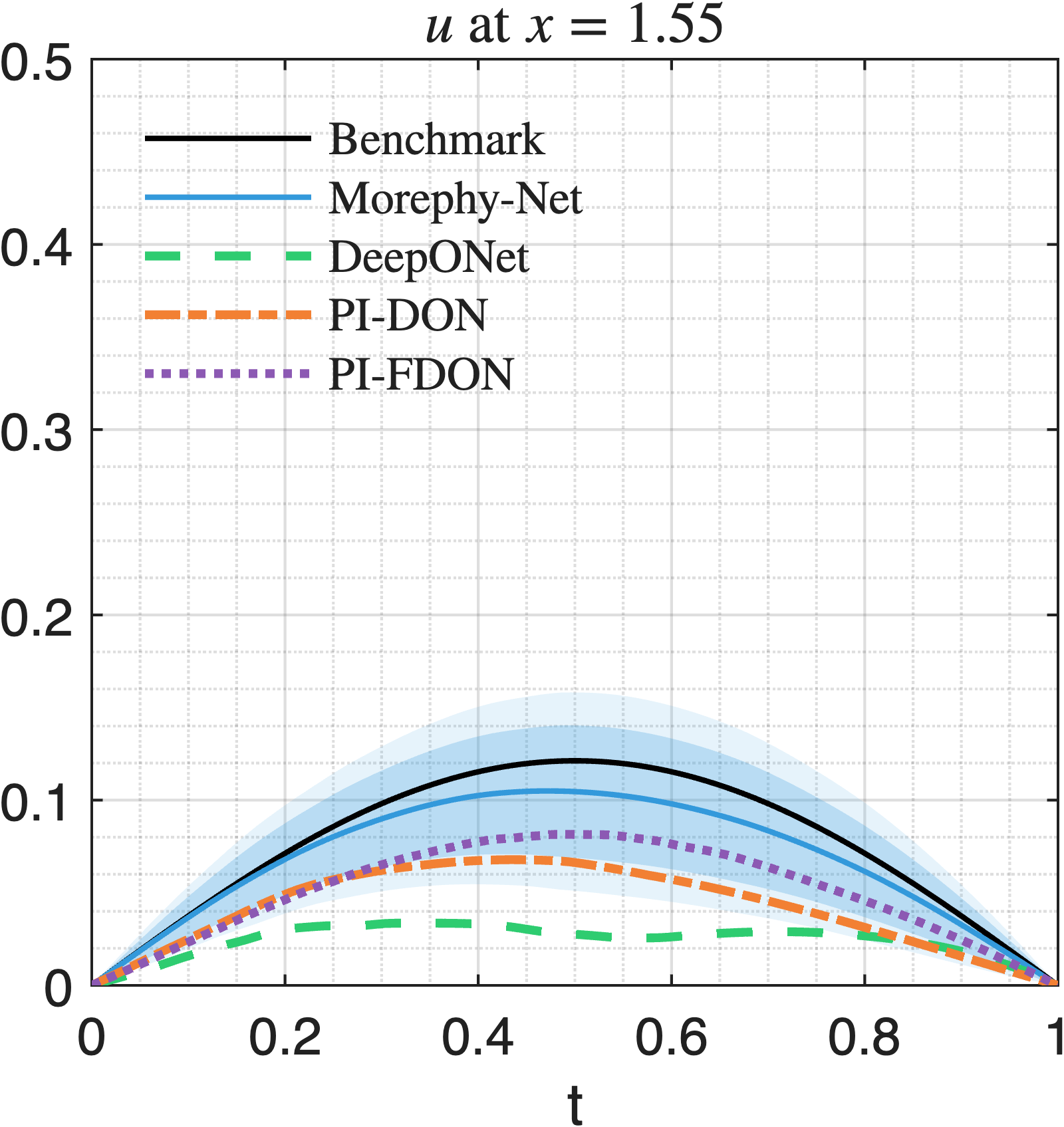}
 \includegraphics[width=.18\linewidth,height=.16\linewidth]{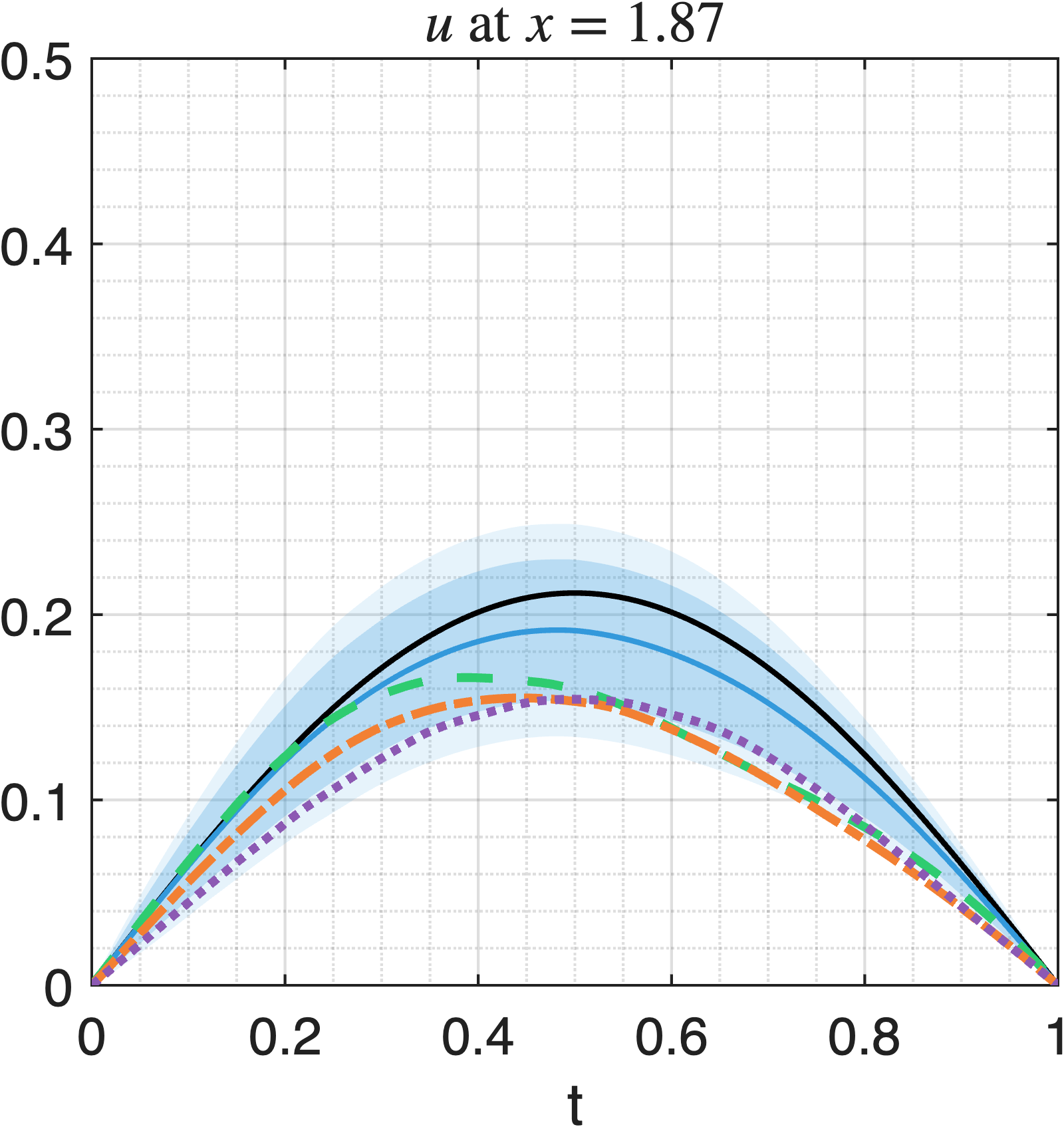}
 \includegraphics[width=.18\linewidth,height=.16\linewidth]{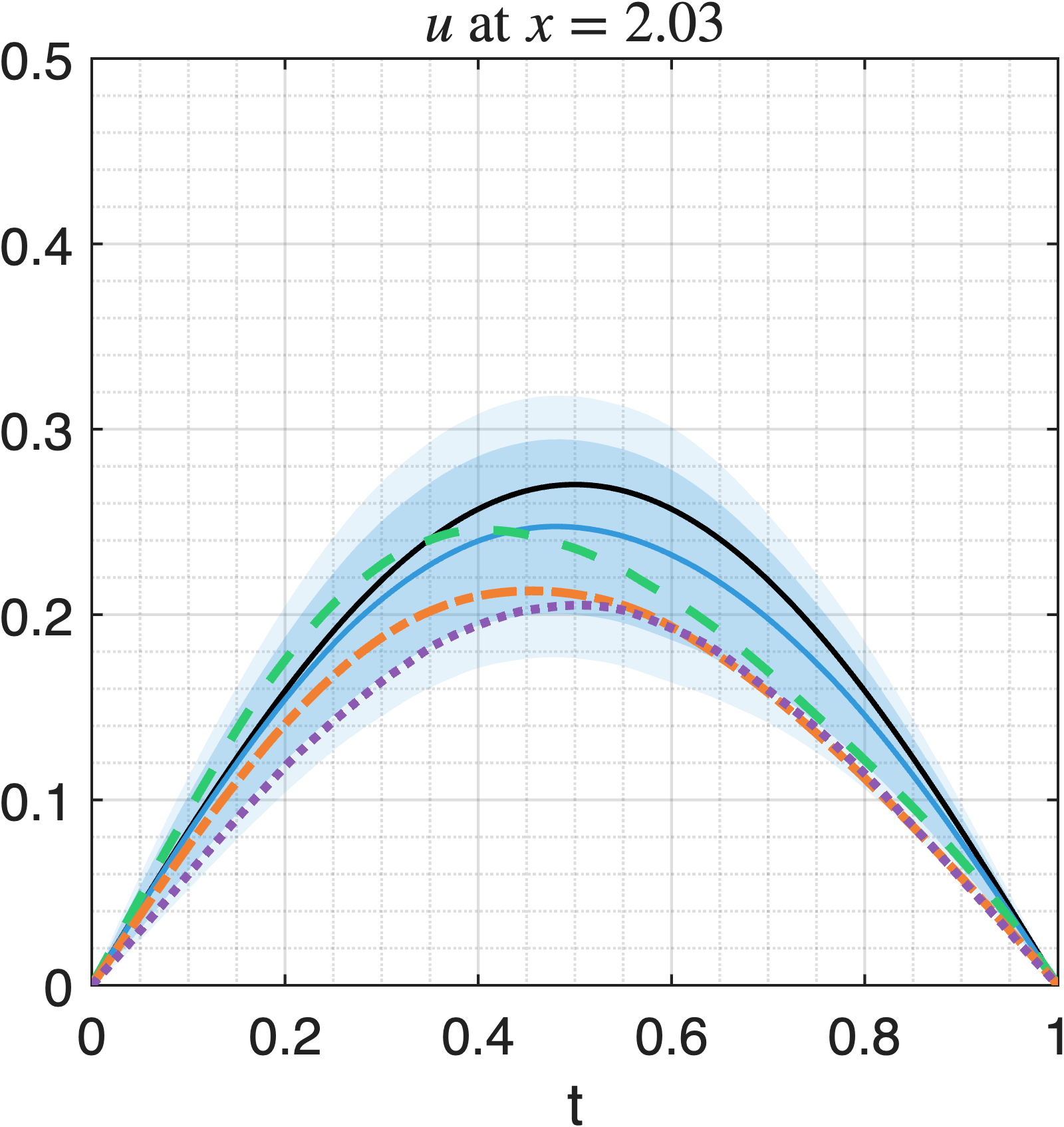}
 \includegraphics[width=.18\linewidth,height=.16\linewidth]{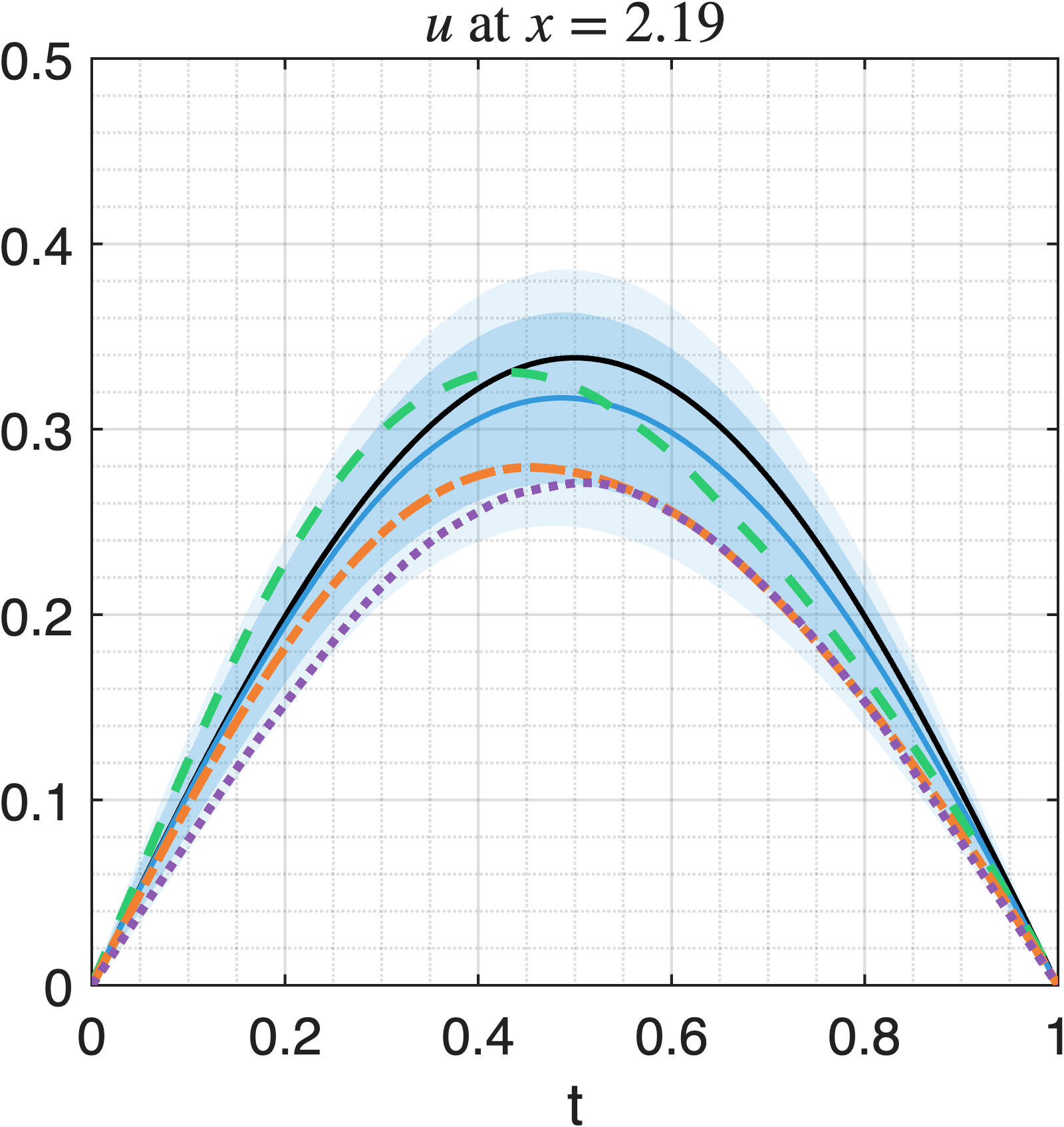}
 \includegraphics[width=.18\linewidth,height=.16\linewidth]{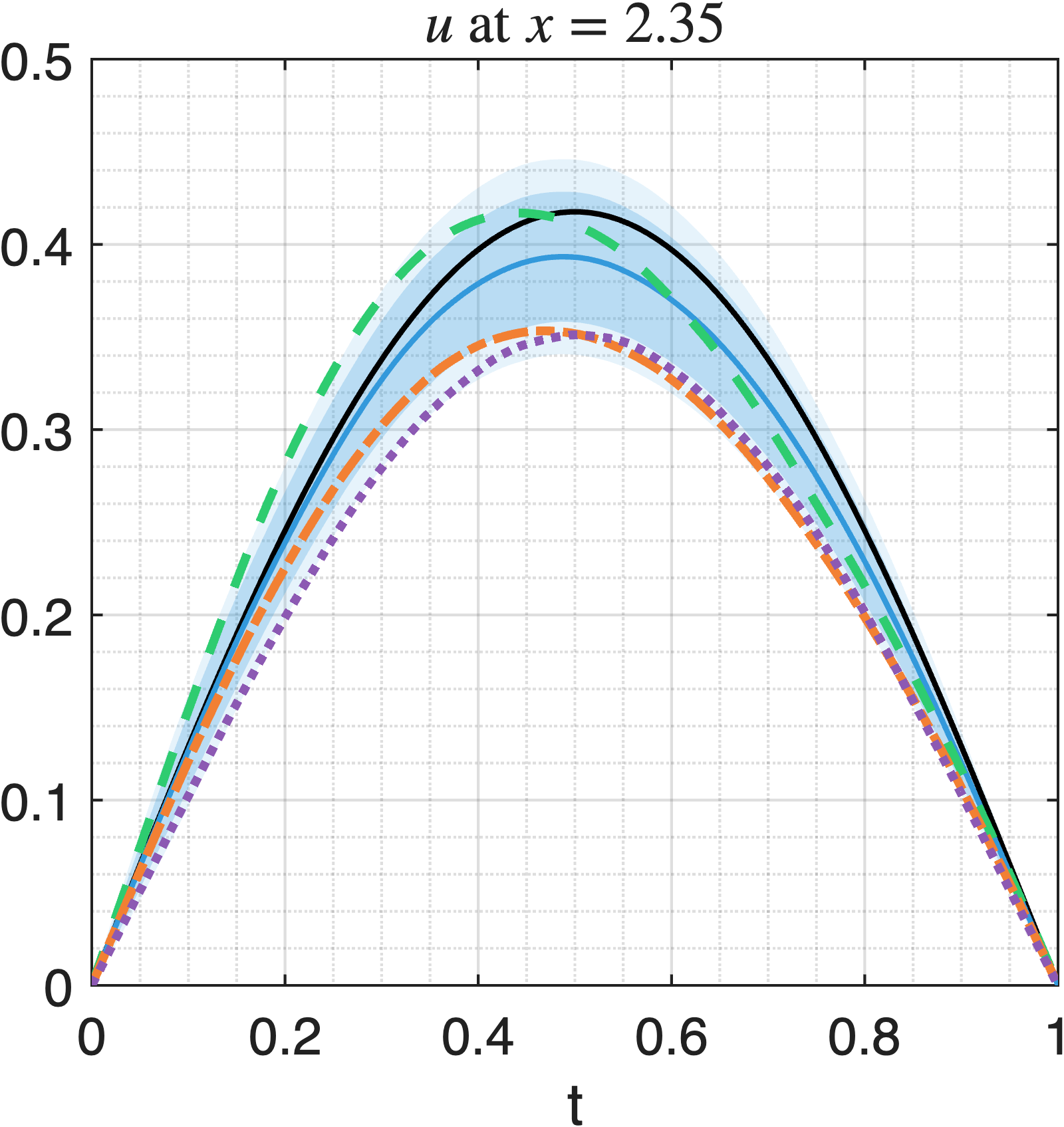}
 }
\\
  \subfloat[Solution $u(x,t)$ at different fixed time $t= t_f$. The blue shading indicates the 95\% confidence interval.     \label{FPDE_forward_UQ_t_result}
]{
 \includegraphics[width=.18\linewidth]{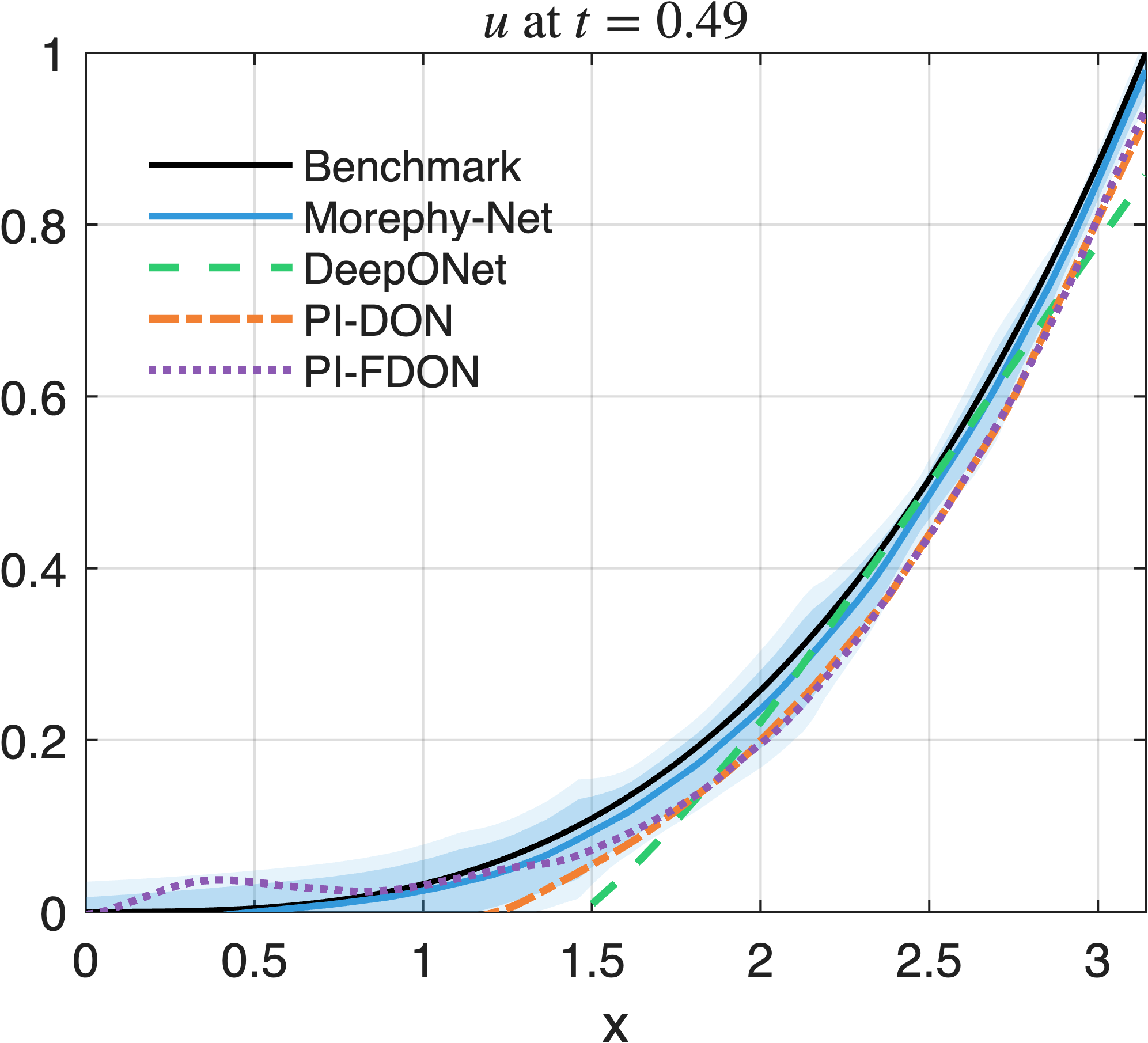}
 \includegraphics[width=.18\linewidth]{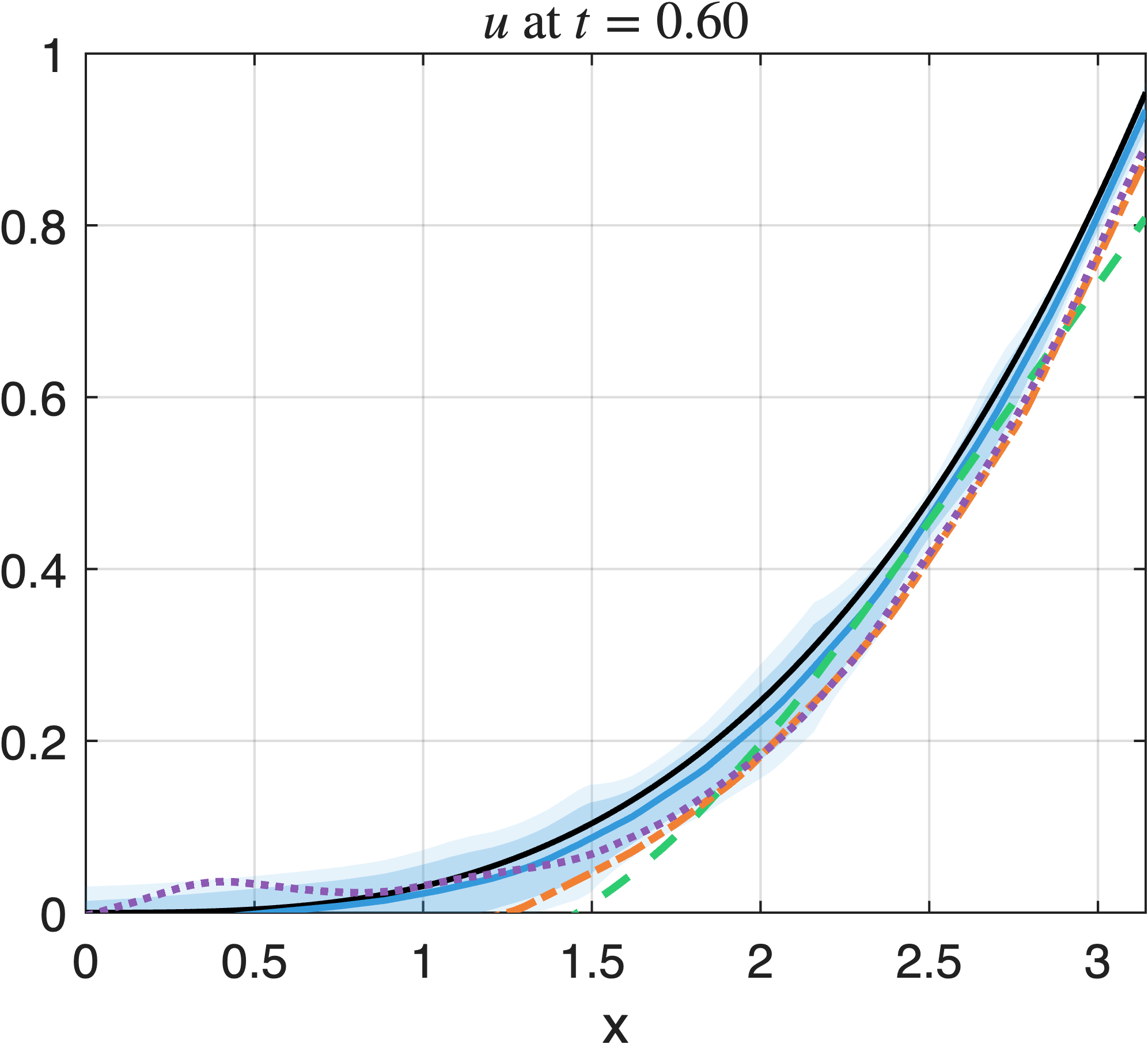}
 \includegraphics[width=.18\linewidth]{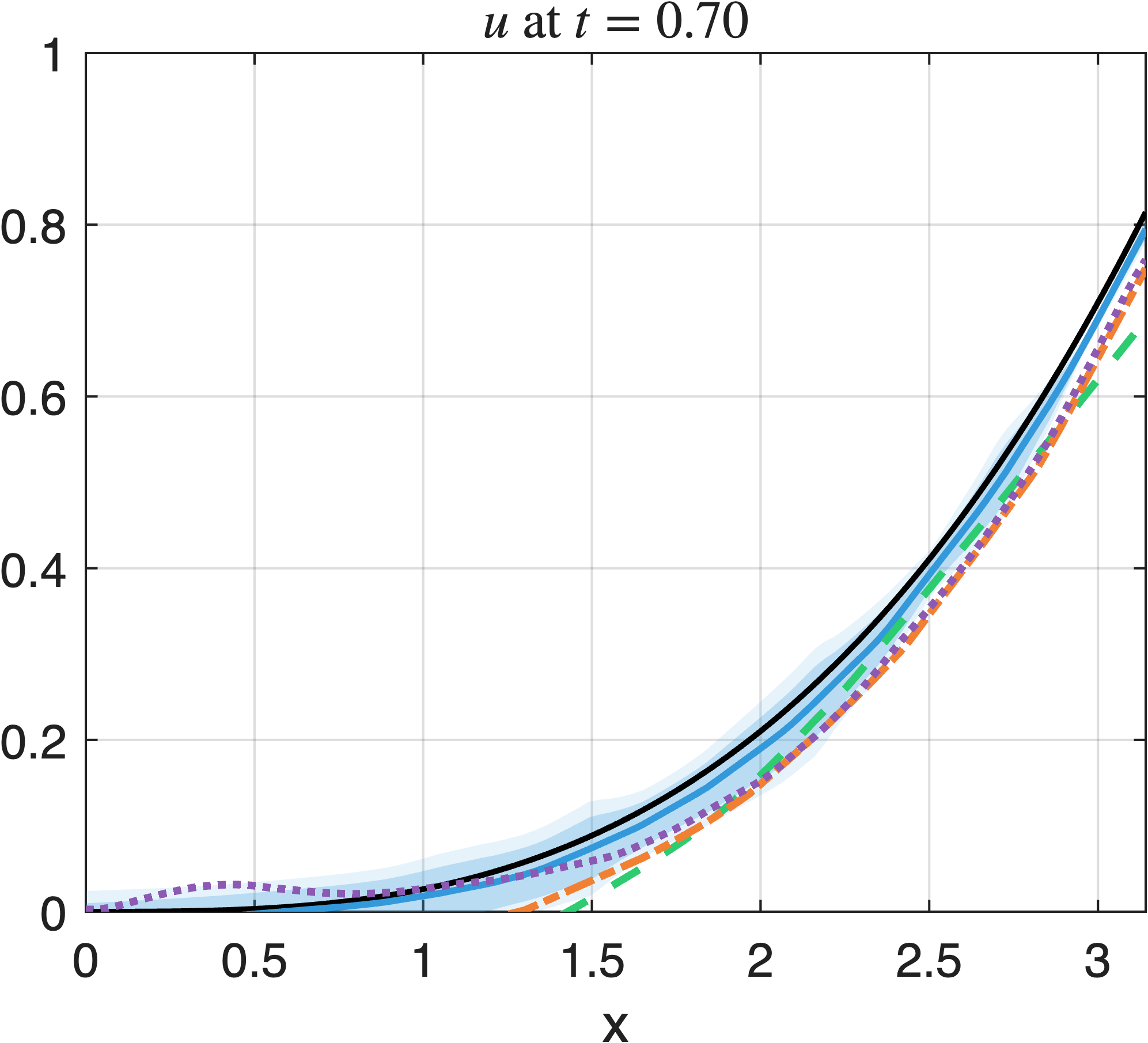}
 \includegraphics[width=.18\linewidth]{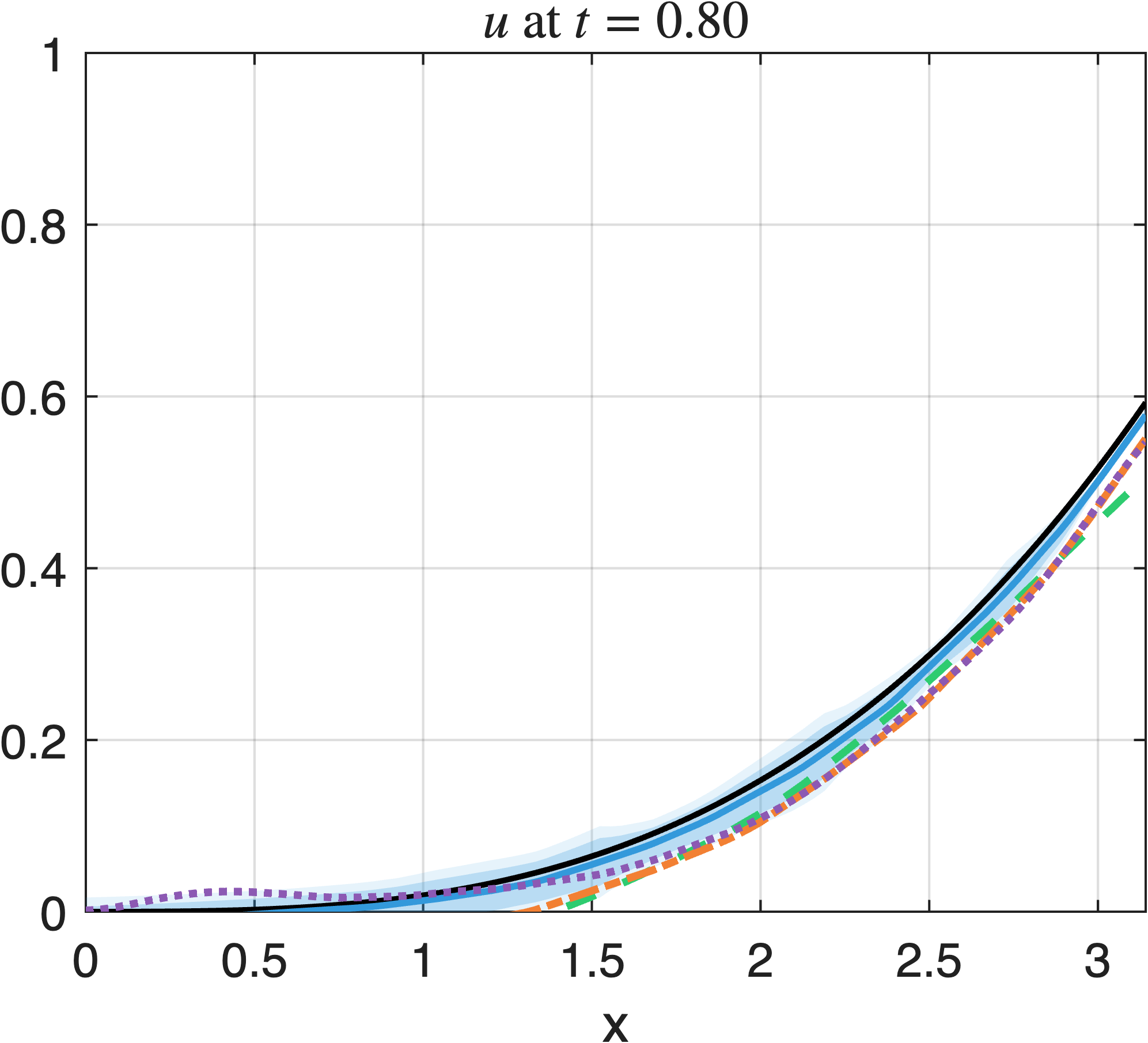}
 \includegraphics[width=.18\linewidth]{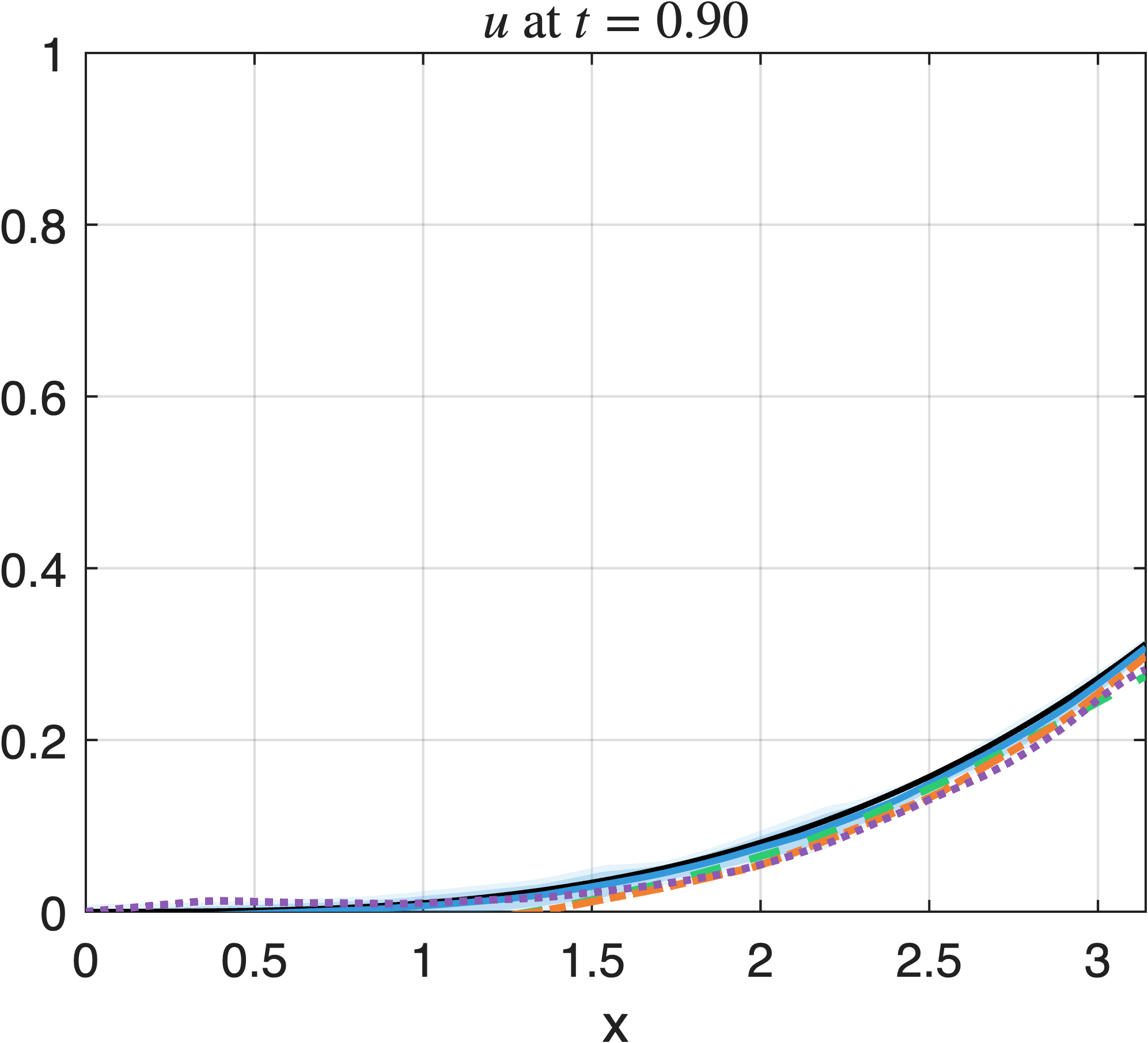}
 }
\\
 \subfloat[$L^1$ Errors     \label{FPDE_forward_result_error}
]{\includegraphics[width=\linewidth]{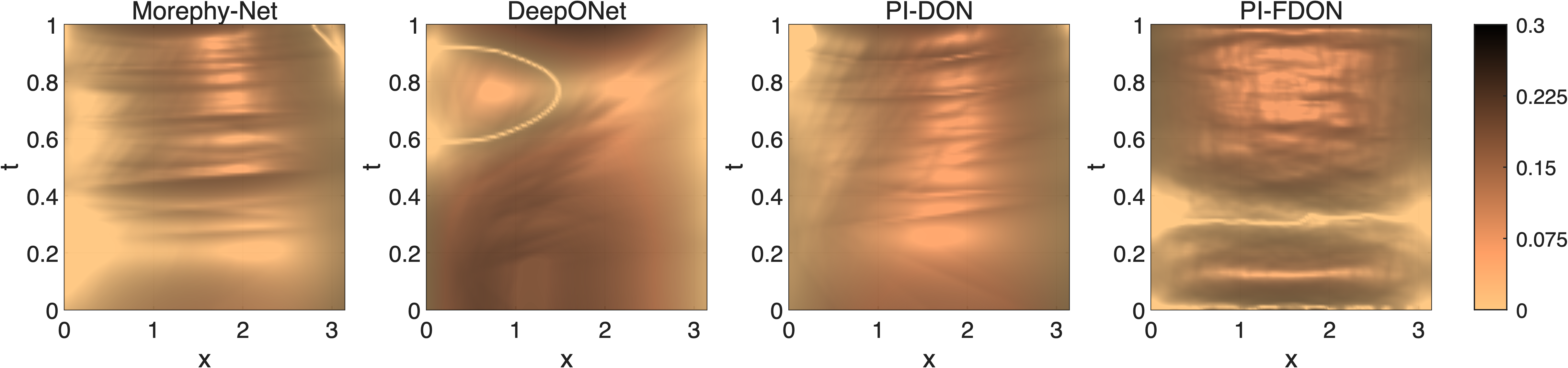}}

    \caption{Comparison of different models for the forward TFMDWEs. The ground truth is compared against predictions from DeepONet, PI-DON, PI-FDON and Morephy-Net. In (b) and (c), the blue shaded regions indicate the corresponding $95\%$ confidence intervals for Morephy-Net.}
    \label{FPDE_forward_result_prediction}
\end{figure*}

This section details the performance evaluation of the four models—Morephy-Net, DeepOnet, PI-DON, and PI-FDON—on the forward problem. A comprehensive comparison against the benchmark solution is presented in Figure~\ref{FPDE_forward_result_prediction}, which includes spatiotemporal field reconstructions, solution cross-sections, and quantitative error distributions. The collective results demonstrate the superior accuracy and reliability of the Morephy-Net architecture. Figure~\ref{FPDE_forward_result} provides a  comparison of the full spatiotemporal solution fields. The predictions generated by Morephy-Net, PI-DON, and PI-FDON are visually indistinguishable from the benchmark, indicating that they have successfully learned the underlying dynamics of the system. In contrast, the DeepOnet model exhibits a noticeable deviation, with a significant error concentrated near the initial condition at $t=0$, suggesting a failure to correctly capture the initial state of the system. Figures~\ref{FPDE_forward_UQ_x_result} and Figure~\ref{FPDE_forward_UQ_t_result} display solution cross-sections at various fixed spatial locations ($x_f$) and temporal snapshots ($t_f$) respectively. The 95\% confidence interval of the Morephy-Net prediction, represented by the blue shaded region, consistently envelops the ground truth benchmark line across all subplots. This demonstrates that the true solution lies within the model's predicted uncertainty bounds. On the contrary, we cab observe the predictions result of other models are lower than the benchmark at fixed location shown in~\ref{FPDE_forward_UQ_x_result} and predilections are deviate from the benchmark shown in~\ref{FPDE_forward_UQ_t_result}.  Figure~\ref{FPDE_forward_result_error} shows the $L^1$ error heatmaps. The error map for Morephy-Net is visibly lighter in color across the entire spatiotemporal domain, indicating a significantly lower overall prediction error compared to the other models.

Table~\ref{FPDE_forward_table_models} demonstrates a consistent trend of improvement across the operator-learning architectures. Morephy-Net stands out as the most accurate model. It achieves the lowest $L^2$ relative error of 0.0935, a substantial improvement over the best-performing operator-learning baseline. It also records the lowest initial and boundary condition (IBC) loss, indicating superior satisfaction of the problem constraints.

\begin{table}[htbp]
\footnotesize
\centering
\caption{Comparison of model performance for the Losses and $L^2$ relative error (RE) in the forward TFMDWEs. }
\label{FPDE_forward_table_models}
\begin{tabular}{l c c c c c }
\toprule
\textbf{Model}& \textbf{Residual loss} & \textbf{BC loss} & \textbf{$L^2$ RE}\\
\midrule
DON      &0.0903& 0.0173    & 0.3080 \\
PI-DON   &0.0703 &0.0153      & 0.1620 \\
PI-FDON  &0.0539 &0.0148     & 0.1319 \\
\textbf{Morephy-Net}   &\textbf{0.0488} &\textbf{0.0121}     & \textbf{0.0955} \\
\bottomrule
\end{tabular}
\end{table}

\subsubsection{Inverse Problem}
\begin{figure*}[!t]
    \centering
 \subfloat[Spatiotemporal fields]{\includegraphics[width=\linewidth]{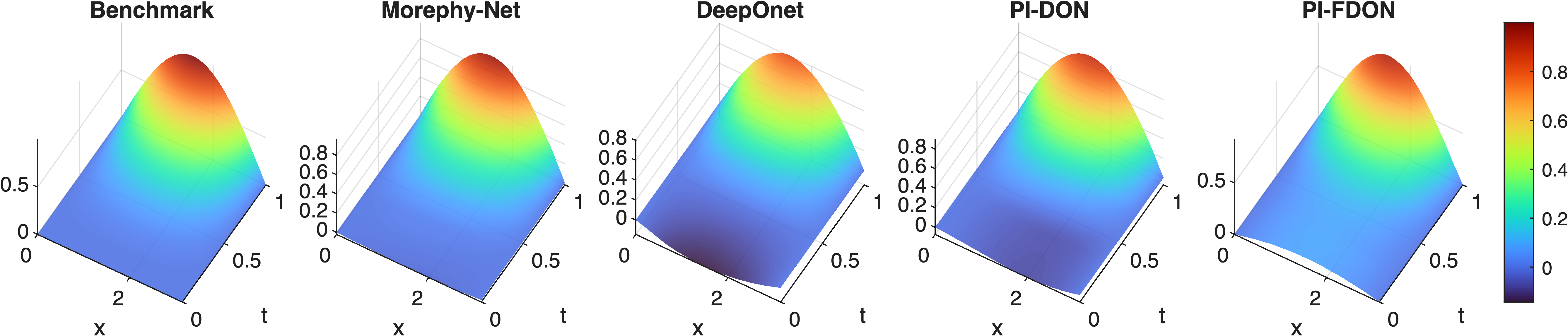}}
 \\
 \subfloat[Solution $u(x,t)$ at different fixed location $x= x_f$. The blue shading indicates the 95\% confidence interval.     \label{FPDE inverse UQ result}
]{
 \includegraphics[width=.18\linewidth,height=.16\linewidth]{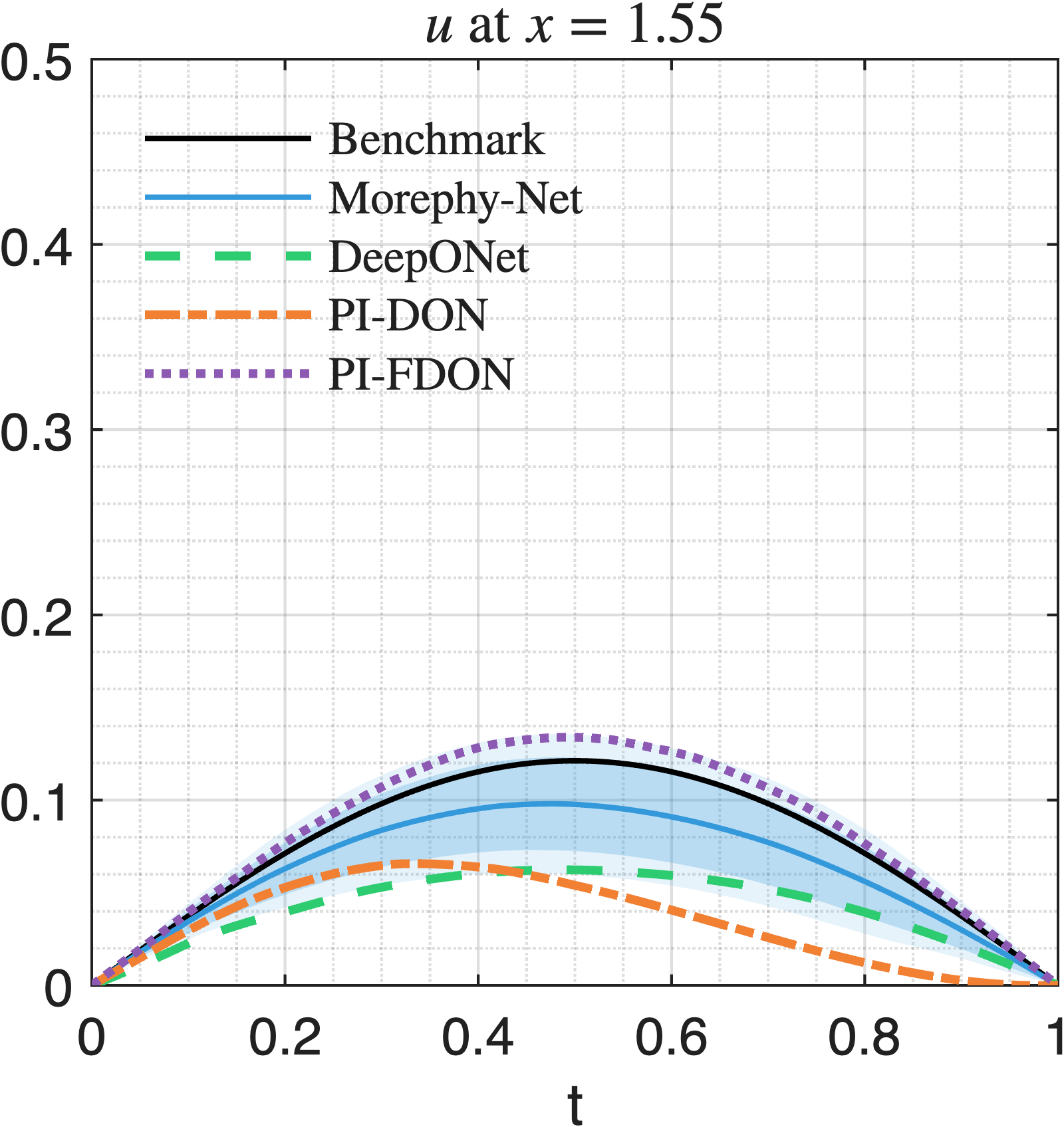}
 \includegraphics[width=.18\linewidth,height=.16\linewidth]{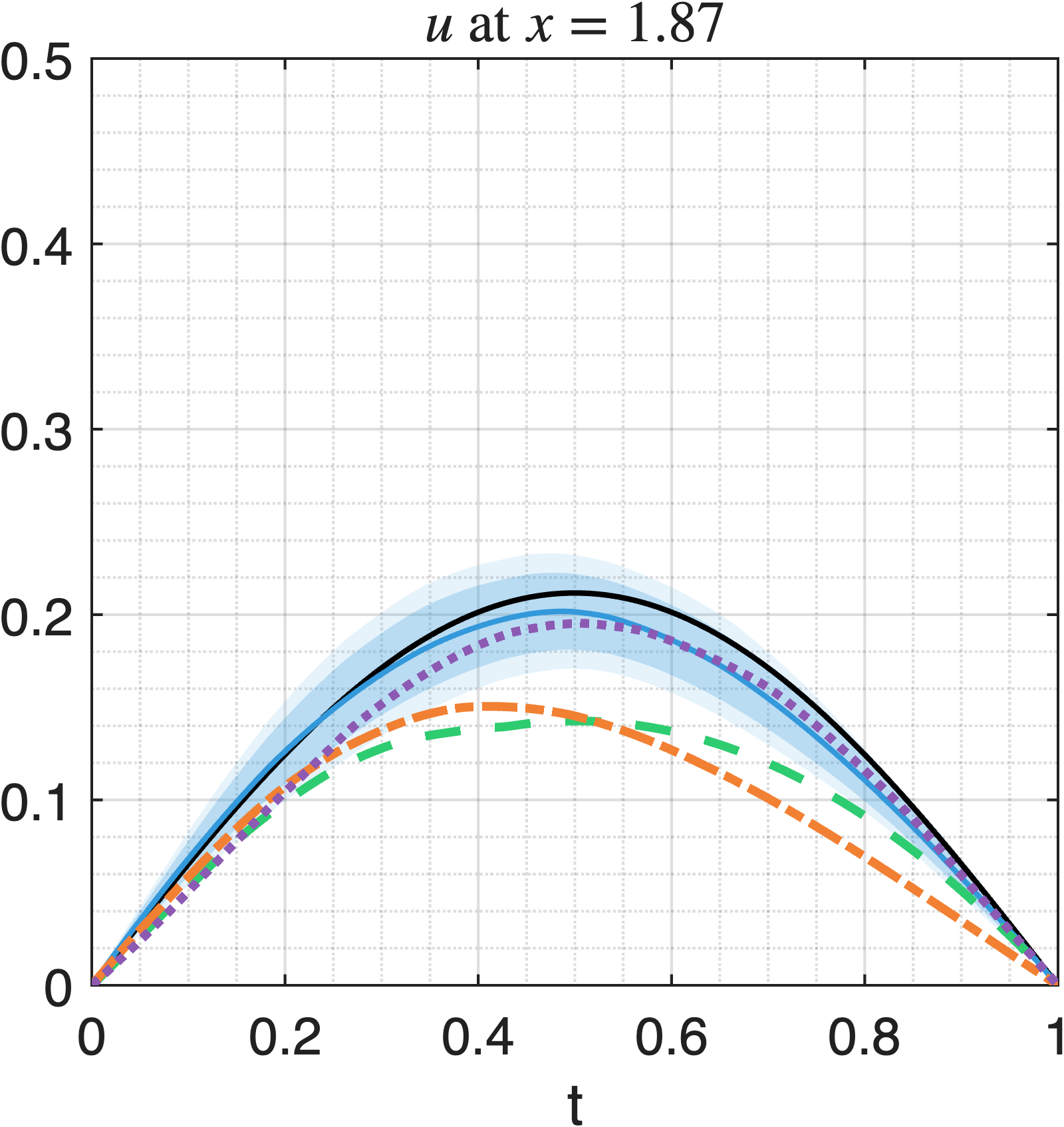}
 \includegraphics[width=.18\linewidth,height=.16\linewidth]{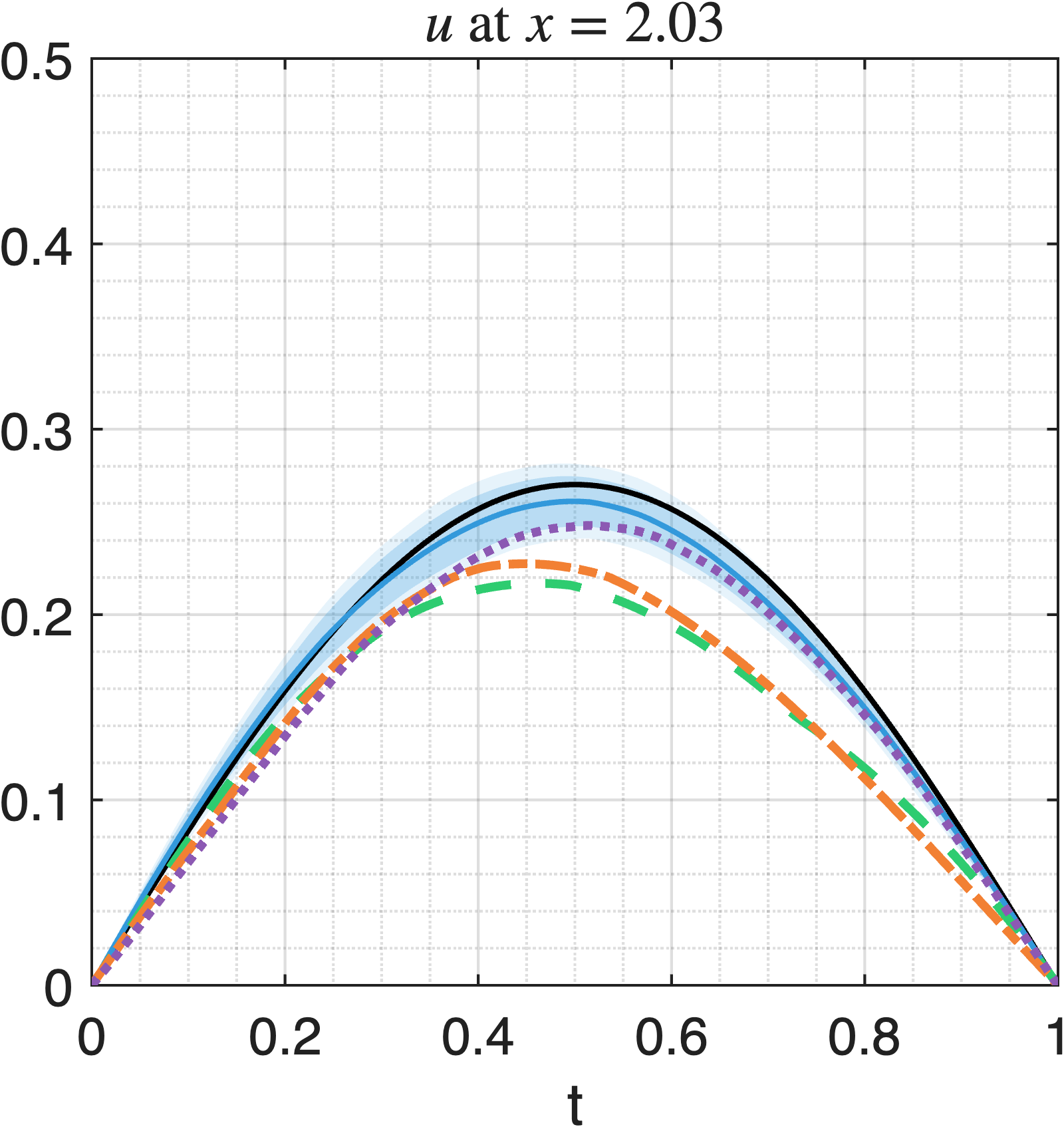}
 \includegraphics[width=.18\linewidth,height=.16\linewidth]{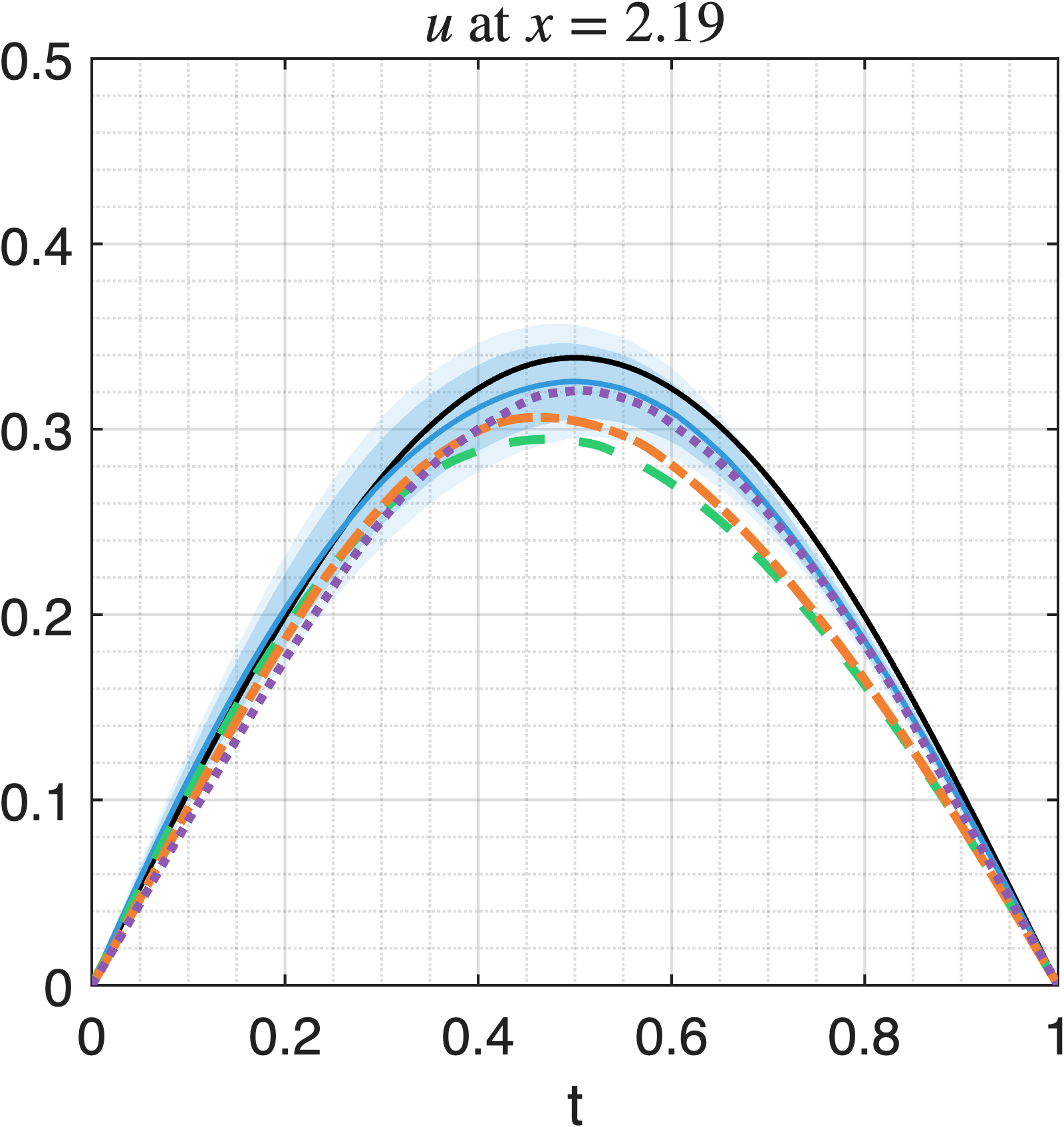}
 \includegraphics[width=.18\linewidth,height=.16\linewidth]{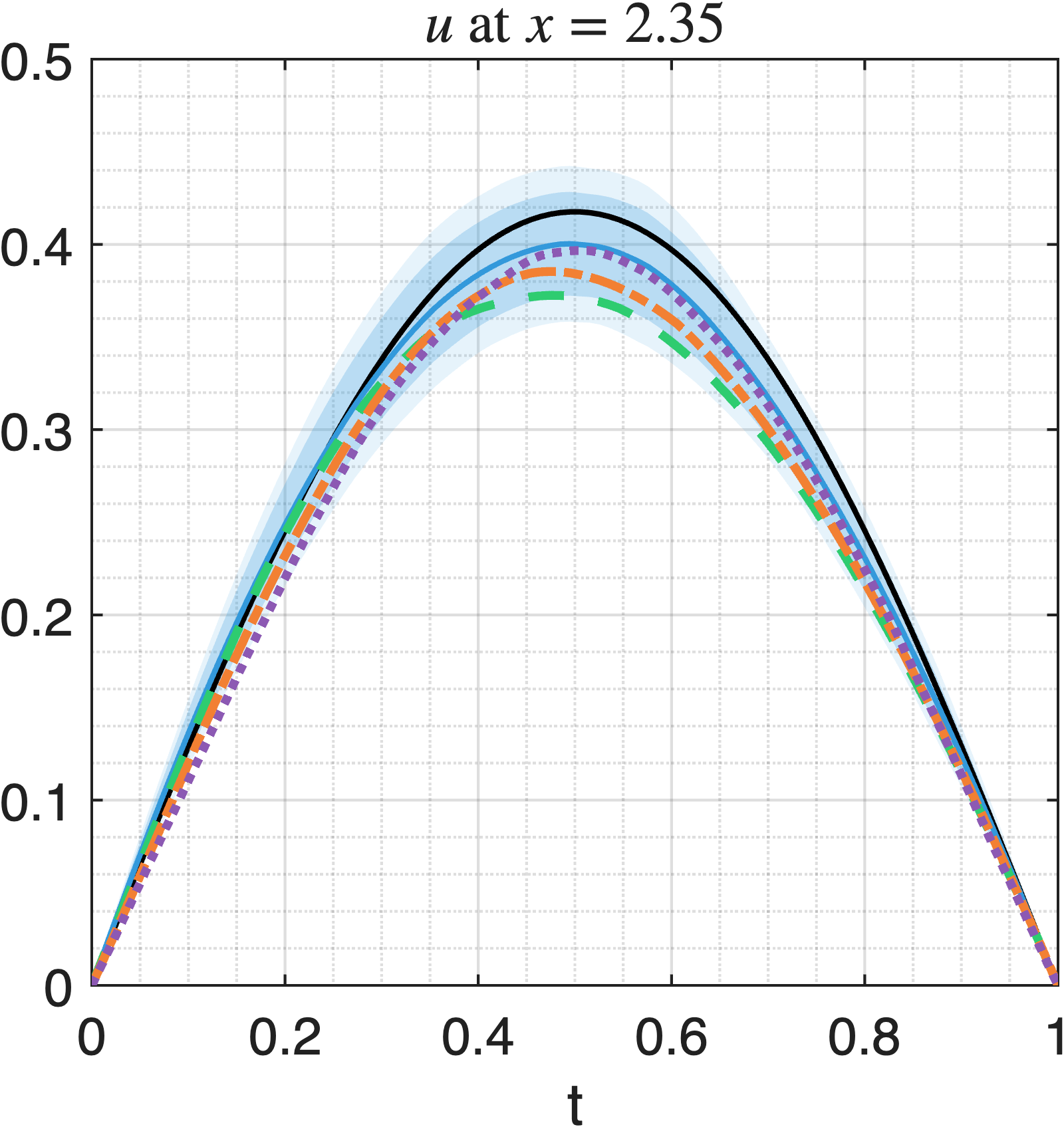}
 }   
 \\
  \subfloat[Solution $u(x,t)$ at different fixed time $t= t_f$. The blue shading indicates the 95\% confidence interval.    \label{FPDE forward UQ result t}
]{
 \includegraphics[width=.18\linewidth]{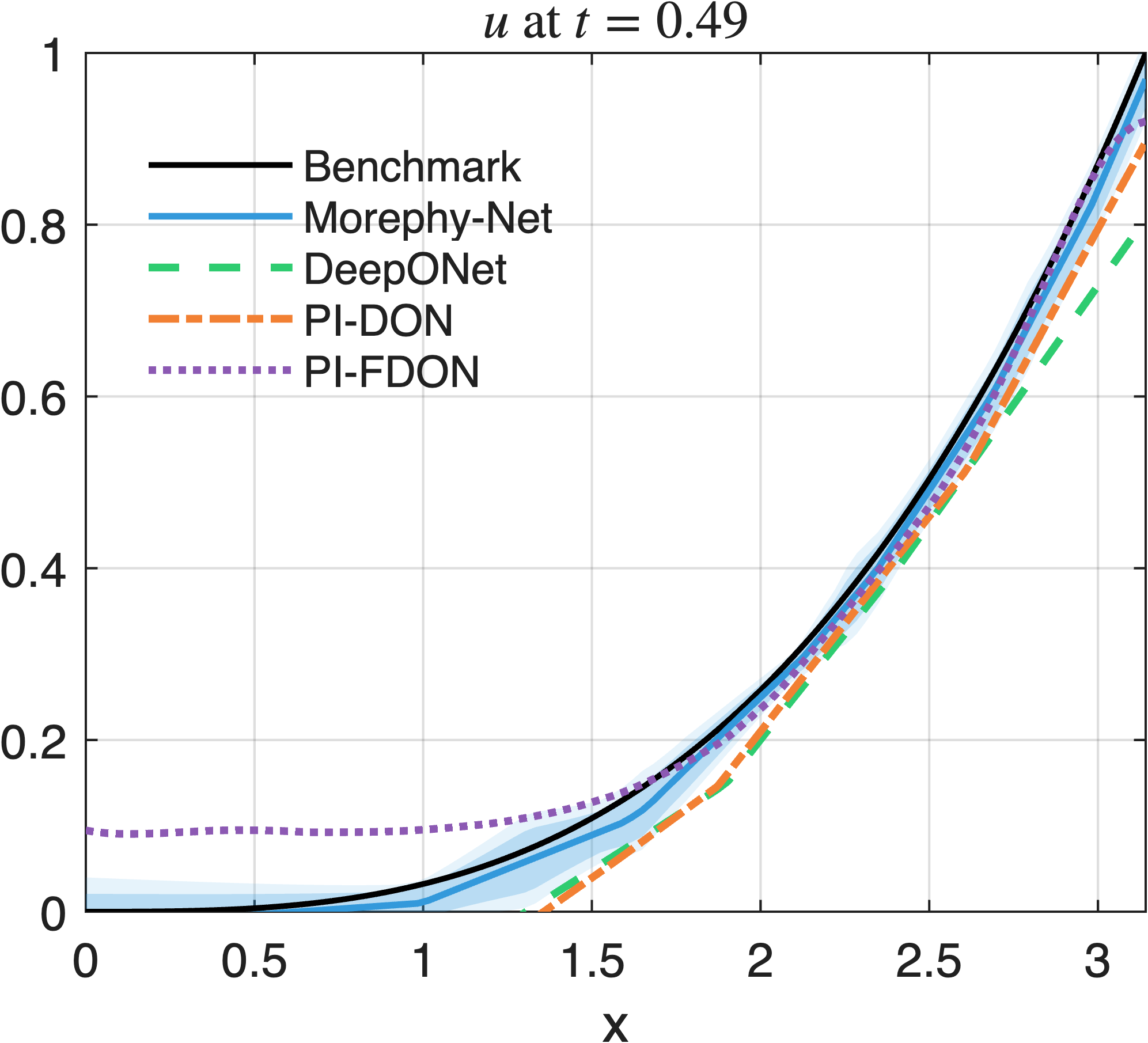}
 \includegraphics[width=.18\linewidth]{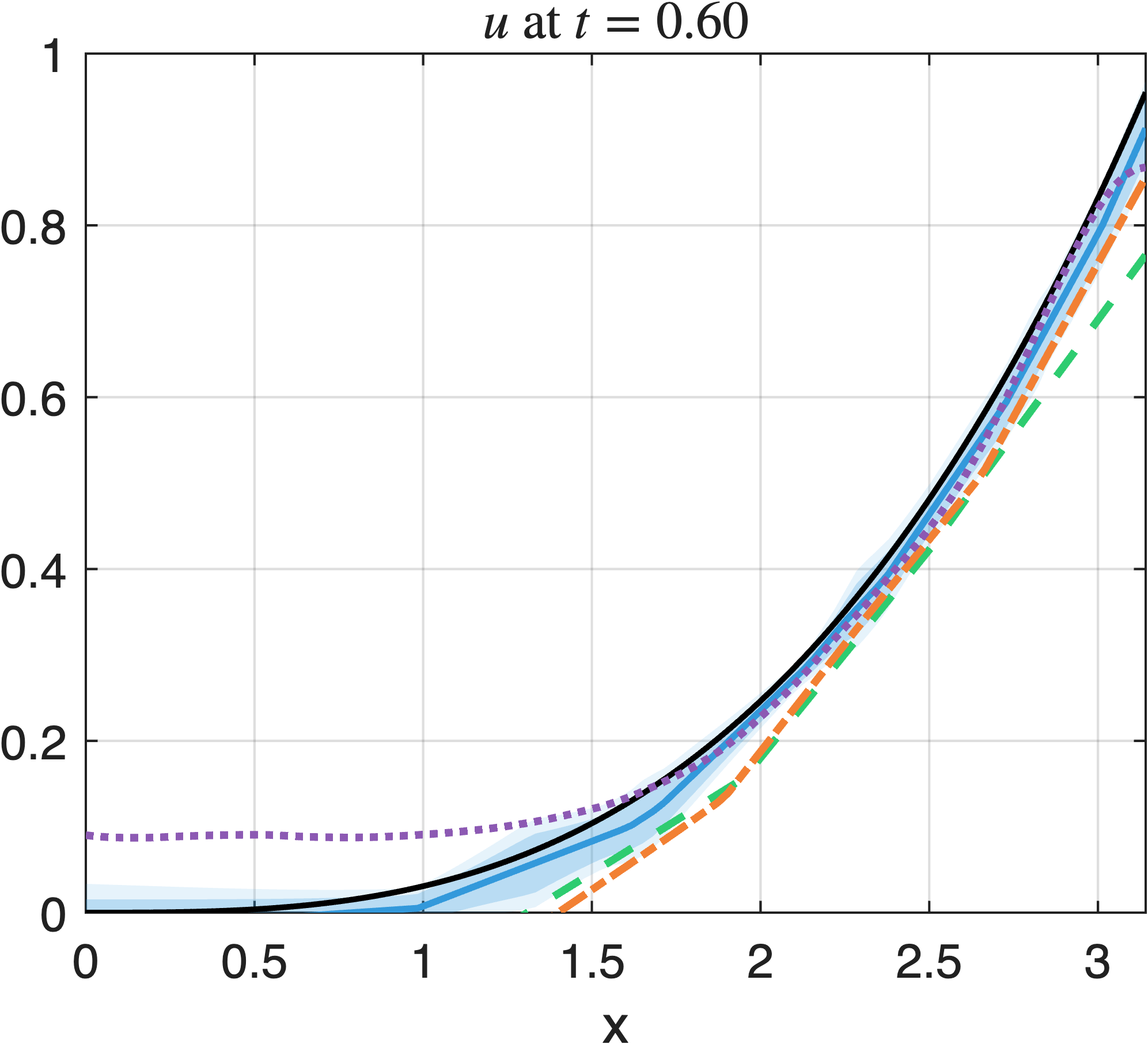}
 \includegraphics[width=.18\linewidth]{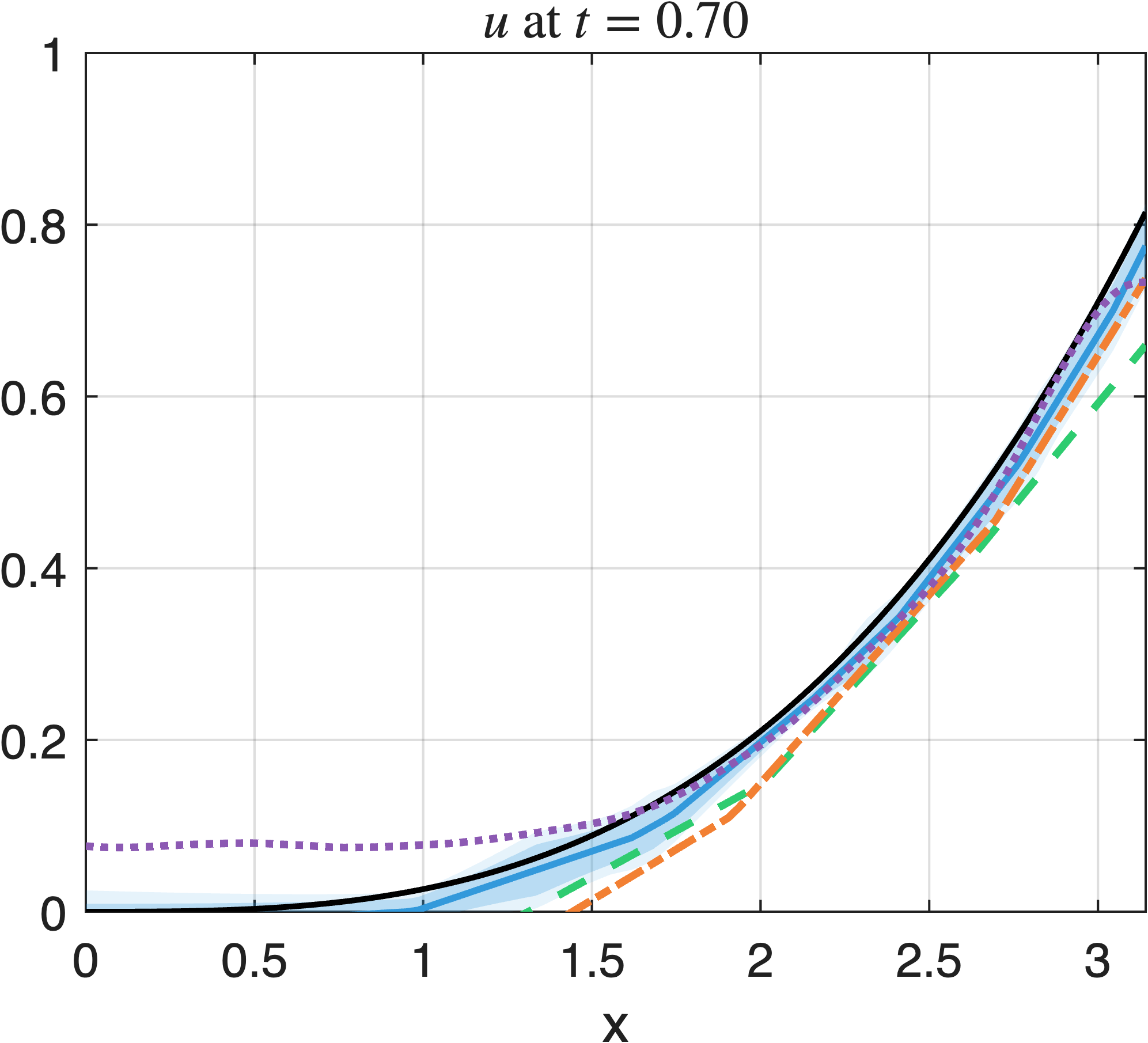}
 \includegraphics[width=.18\linewidth]{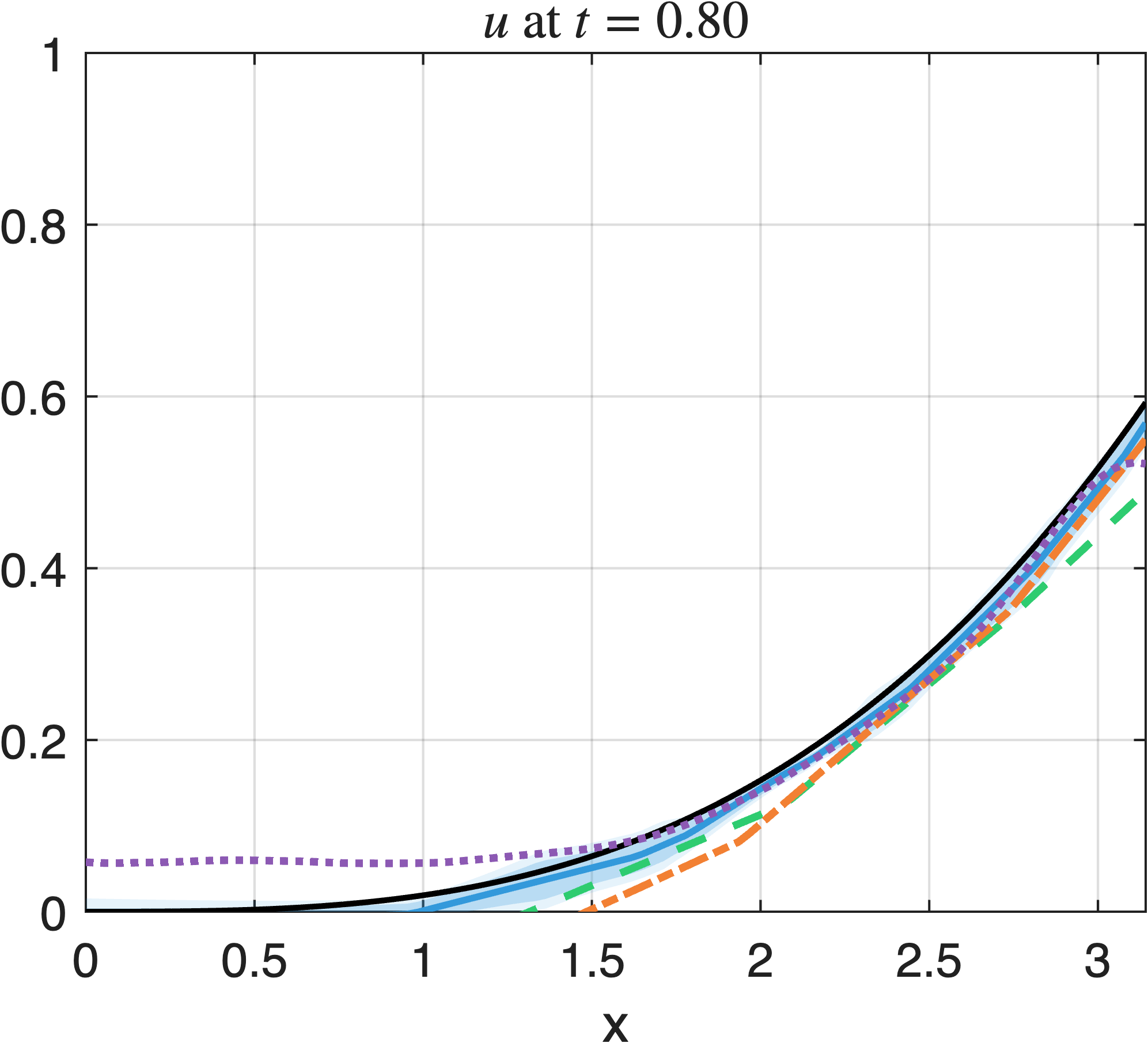}
 \includegraphics[width=.18\linewidth]{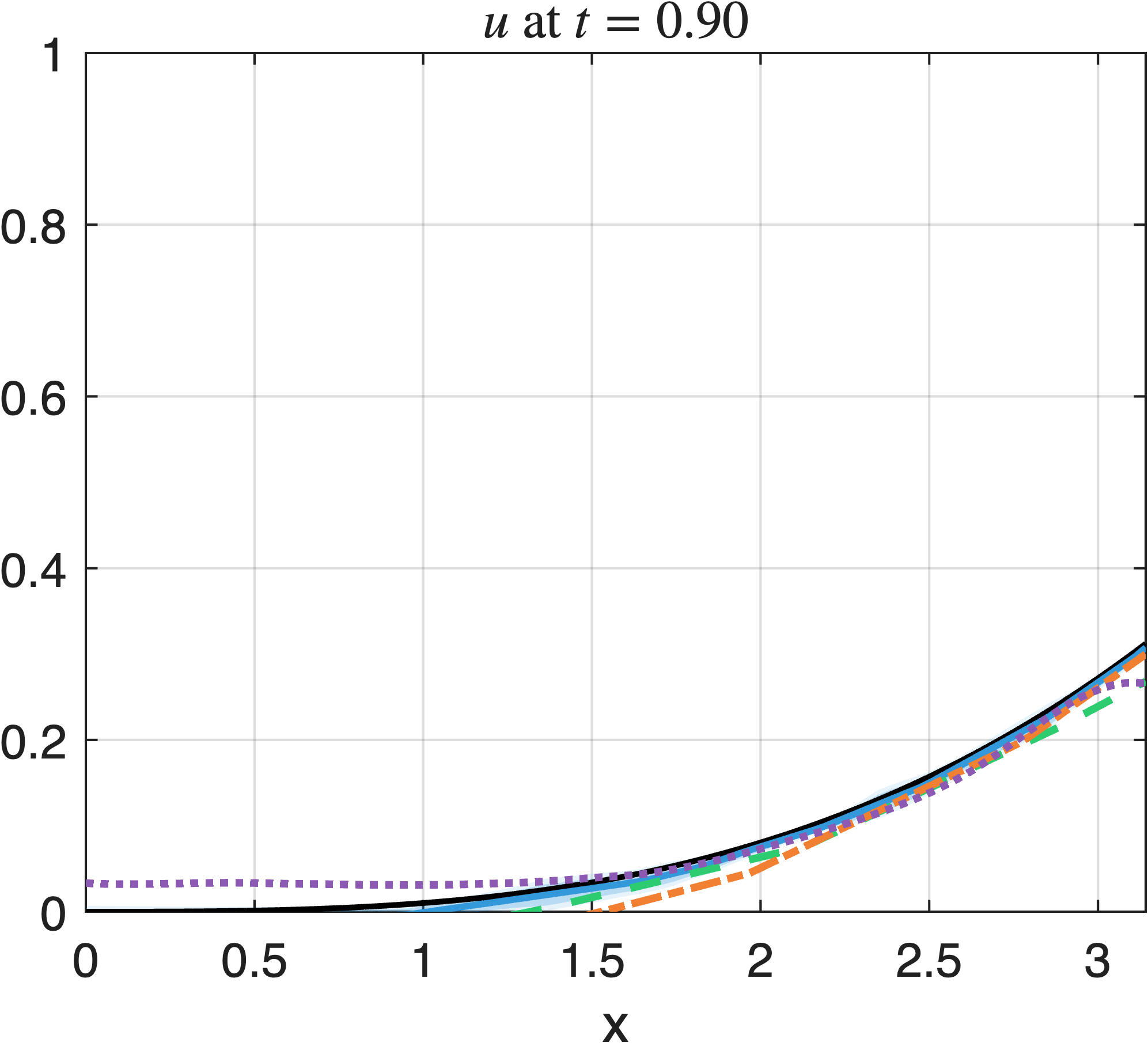}
 }
 \\
  \subfloat[$L^1$ Errors         \label{FPDE inverse result error}
]{\includegraphics[width=\linewidth]{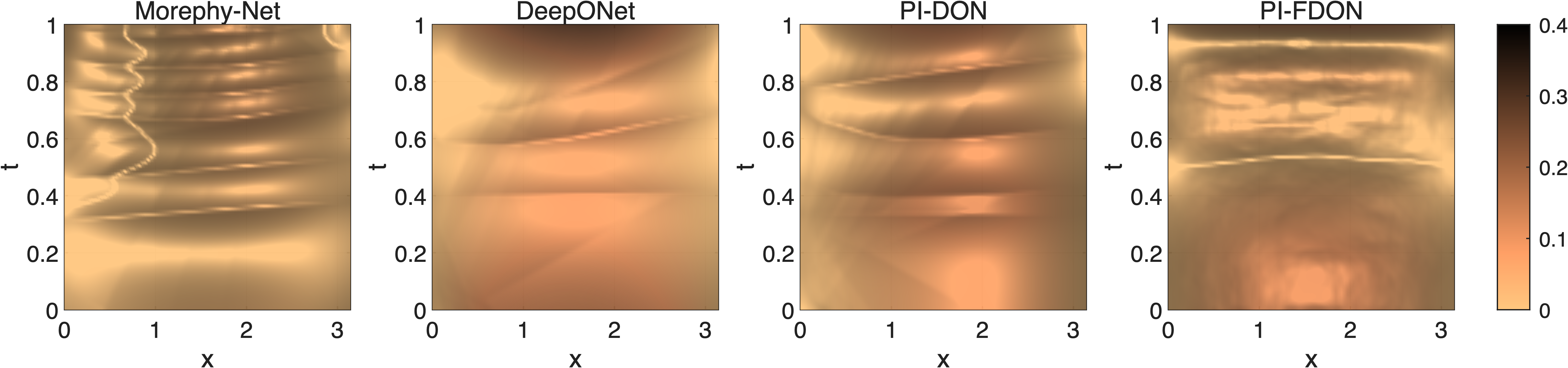}}
    \caption{Comparison of different models for the inverse TFMDWEs. The ground truth is compared against predictions from DeepONet, PI-DON, PI-FDON and Morephy-Net. In (b) and (c), the blue shaded regions indicate the corresponding $95\%$ confidence intervals for Morephy-Net.}
    \label{FPDE_inverse_result_prediction}
\end{figure*}

We consider the inverse TFMDWE setting and train the model to infer the fractional order and reconstruct the solution. Figure~\ref{FPDE_inverse_result_prediction} visualizes the results. In the first row, we compare the reconstructed spatiotemporal fields from different models with the benchmark. Morephy-Net matches the benchmark closely, whereas the other three models exhibit noticeable mismatches near $x=0$ and $t=0$. The second and third rows show 2D projections of the solution at different fixed locations and times. These plots indicate that Morephy-Net consistently covers the benchmark within its 95\% confidence interval (dark-blue shading). The fourth row shows the $L^1$ error heatmaps for the four models, where Morephy-Net exhibits smaller errors overall.

\begin{table*}[htbp]
\footnotesize
\centering
\caption{Comparison of model performance for the Losses and $L^2$ relative error (RE) in the inverse TFMDWEs}
\label{tab:FPDE Inverse Problem model}
\begin{tabular}{l c c c c c  c}
\toprule
\textbf{Model} &\textbf{Residual Loss}&\textbf{BC Loss}&\textbf{Data Loss}& \textbf{Predicted $\alpha$}  & \textbf{$L^2$ RE}\\
\midrule
DON     &0.0641 & 0.0275&0.0503 & 0.6930        & 0.2298 \\
PI-DON   &0.0541&0.0148 & 0.0499& 0.6640       & 0.1813 \\
PI-FDON  &0.0509&0.0151 &0.0458 & 0.5954       & 0.1359 \\
\textbf{Morephy-Net}   &\textbf{0.0236} &\textbf{0.0148}& \textbf{0.0384} & \textbf{0.5740}     & \textbf{0.1134} \\
\bottomrule
\end{tabular}
\end{table*}

\section{Conclusion and Future Work \label{sec:conclusion}}
In this paper, we introduced Morephy-Net, a physics-informed operator-learning framework that integrates evolutionary multi-objective optimization with replica-exchange stochastic sampling. By treating data/operator and physics residual losses as separate objectives and searching the Pareto front, Morephy-Net removes the need for hand-tuned loss weights. Meanwhile, reSGLD improves global exploration of the parameter space and enables principled Bayesian uncertainty quantification. Numerical experiments on the Burgers' equation and the time-fractional mixed diffusion--wave equation demonstrate that Morephy-Net achieves higher accuracy and stronger robustness to noisy observations than standard operator-learning baselines (e.g., DeepONet, PI-DON, and PI-FDON), in both forward and inverse settings.
In addition to improved point estimates, Morephy-Net provides calibrated uncertainty information---a capability that is typically absent in deterministic operator-learning pipelines---which is critical for reliable scientific inference and decision-making.

There are several promising future directions. First, extending the proposed multi-objective replica-exchange physics-informed setting to evolutionary Kolmogorov--Arnold networks \cite{cruz2025state,lin2025energy} may improve both expressiveness and interpretability. Second, Morephy-Net's inherent uncertainty quantification (UQ) makes it a natural candidate for data assimilation (DA), where accurate state estimation under model uncertainty is crucial for high-dimensional dynamical systems \cite{qi2025data,law2015data,chen2020efficient,durant2025updating,liu2023statistical,wang2016multi}. Since DA often requires many repeated forecasts in an ensemble or iterative framework, the operator-learning structure of Morephy-Net can serve as a computationally efficient surrogate for the underlying dynamics; combined with UQ, it is particularly well-suited for real-time assimilation with noisy observations.
Third, applying Morephy-Net to turbulent flows governed by the Navier–Stokes equations (NSE) offers an exciting opportunity to test its robustness and scalability in highly nonlinear and multiscale systems \cite{mou2023energy,heinonen2020turbulence,yang2012adaptive}. 

\section*{Github codes} 
The source code for Morephy-Net is available at: \url{https://github.com/jeremylu916/MORPHY-Net}

\section*{Acknowledgment}
We would like to thank the support of National Science Foundation (DMS-2533878, DMS-2053746, DMS-2134209, ECCS-2328241, CBET-2347401 and OAC-2311848), and U.S.~Department of Energy (DOE) Office of Science Advanced Scientific Computing Research program DE-SC0023161, the SciDAC LEADS Institute, and DOE–Fusion Energy Science, under grant number: DE-SC0024583.

\bibliographystyle{unsrt}  
\bibliography{ref}
\end{document}